\documentclass[10pt,twocolumn]{article}

\usepackage[margin=0.72in,columnsep=0.25in]{geometry}
\usepackage[T1]{fontenc}
\usepackage[utf8]{inputenc}
\usepackage{lmodern}
\usepackage{microtype}
\usepackage{xcolor}
\usepackage{graphicx}
\usepackage{tikz}
\usetikzlibrary{arrows.meta,positioning,fit,calc,shadows.blur,backgrounds,matrix}
\usepackage[most]{tcolorbox}
\usepackage{booktabs}
\usepackage{tabularx}
\usepackage{array}
\usepackage{amsmath,amssymb}
\usepackage{fancyhdr}
\usepackage{titlesec}
\usepackage{enumitem}
\usepackage{stfloats}
\usepackage{xurl}
\usepackage[hidelinks]{hyperref}
\usepackage[numbers,sort&compress]{natbib}
\definecolor{StateNavy}{HTML}{111111}
\definecolor{StateInk}{HTML}{101828}
\definecolor{StateSlate}{HTML}{344054}
\definecolor{StateBlue}{HTML}{111111}
\definecolor{StateCyan}{HTML}{16FFBB}
\definecolor{StateMint}{HTML}{16FFBB}
\definecolor{StateLime}{HTML}{16FFBB}
\definecolor{StateViolet}{HTML}{111111}
\definecolor{StateLine}{HTML}{D0D5DD}
\definecolor{StateSoft}{HTML}{F7F8FA}
\definecolor{StateWarn}{HTML}{B99A89}
\definecolor{StateRisk}{HTML}{B99A89}
\definecolor{StateShadow}{HTML}{9E9691}
\definecolor{StatePaper}{HTML}{FFFFFF}

\hypersetup{
  colorlinks=true,
  linkcolor=StateBlue,
  citecolor=StateBlue,
  urlcolor=StateViolet
}

\setlist[itemize]{leftmargin=1.25em,itemsep=0.22em,topsep=0.35em}
\setlist[enumerate]{leftmargin=1.35em,itemsep=0.25em,topsep=0.35em}

\newsavebox{\stateeqbox}
\newcommand{\eqfit}[1]{%
  \sbox{\stateeqbox}{$\displaystyle #1$}%
  \ifdim\wd\stateeqbox>0.88\linewidth
    \resizebox{0.88\linewidth}{!}{$\displaystyle #1$}%
  \else
    \usebox{\stateeqbox}%
  \fi
}

\titleformat{\section}{\Large\bfseries\raggedright\color{StateNavy}}{\thesection}{0.6em}{}
\titleformat{\subsection}{\large\bfseries\raggedright\color{StateInk}}{\thesubsection}{0.6em}{}
\titleformat{\subsubsection}{\normalsize\bfseries\raggedright\color{StateNavy}}{\thesubsubsection}{0.5em}{}

\newcommand{\StateTitleLogo}{%
  \IfFileExists{STATE16-LOGO.jpg}{\includegraphics[width=1.42in]{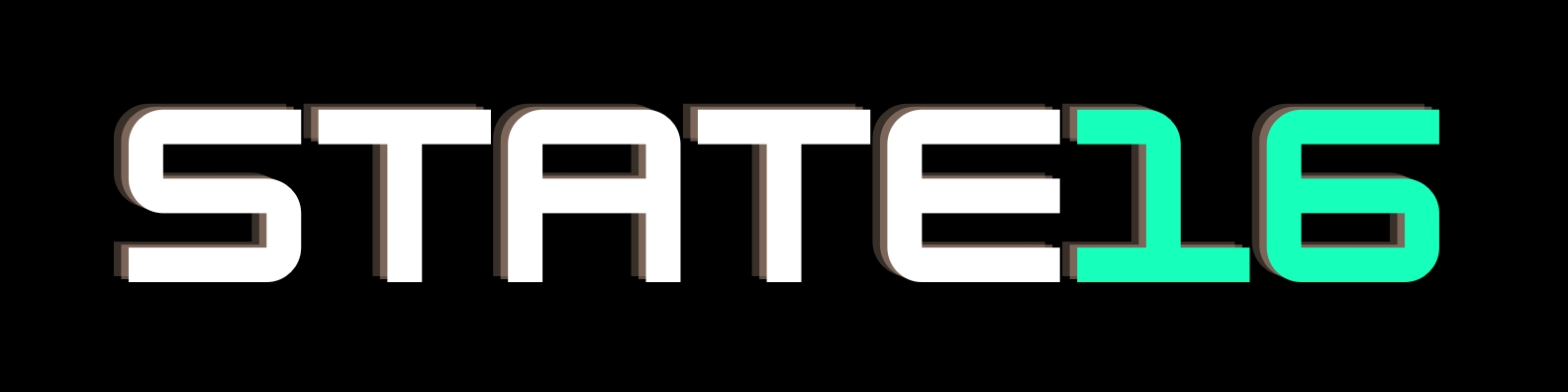}}{%
  \IfFileExists{state16-wordmark-transparent.png}{\includegraphics[width=1.42in]{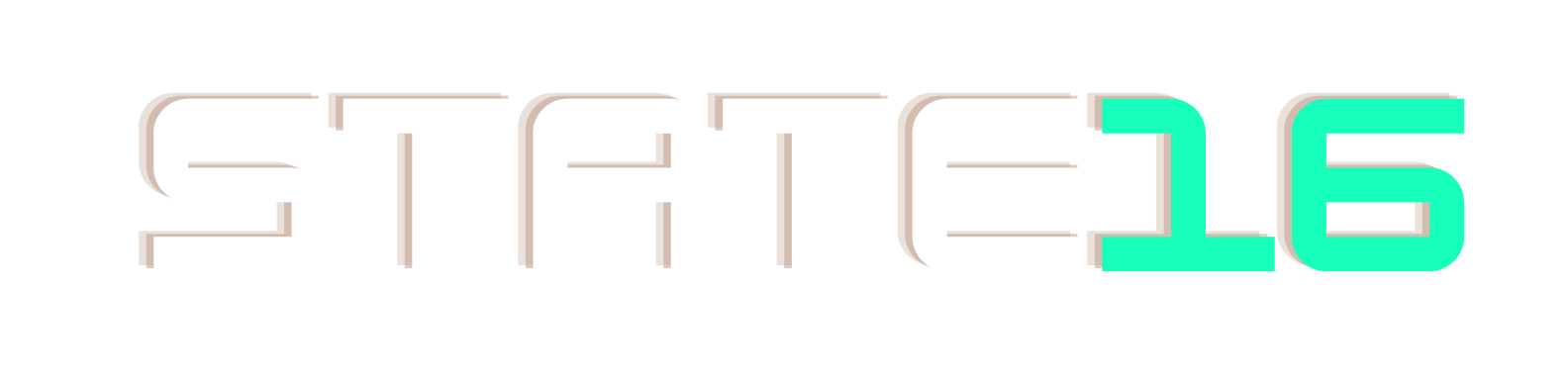}}{%
  \IfFileExists{state16-wordmark.png}{\includegraphics[width=1.42in]{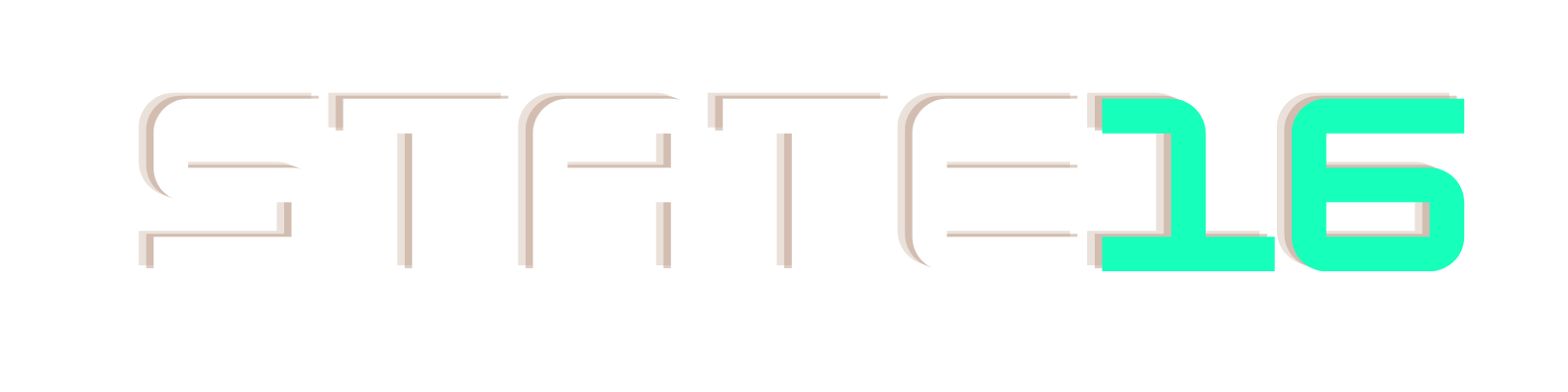}}{%
  \IfFileExists{state16-icon.jpg}{\includegraphics[width=0.68in]{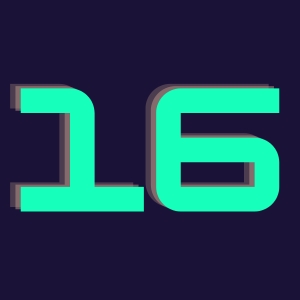}}{}}}}%
}

\newcommand{\StateHeaderLogo}{%
  \IfFileExists{STATE16-LOGO.jpg}{\includegraphics[height=0.14in]{STATE16-LOGO.jpg}}{%
  \IfFileExists{state16-wordmark-transparent.png}{\includegraphics[height=0.14in]{state16-wordmark-transparent.png}}{%
  \IfFileExists{state16-wordmark.png}{\includegraphics[height=0.14in]{state16-wordmark.png}}{%
  \textcolor{StateNavy}{STATE16}}}}%
}

\renewcommand{\headrulewidth}{0.4pt}
\renewcommand{\headrule}{\hbox to\headwidth{\color{StateLine}\leaders\hrule height \headrulewidth\hfill}}

\newcolumntype{Y}{>{\raggedright\arraybackslash}X}

\newcommand{\StateSixteenLogo}{%
  \IfFileExists{STATE16-LOGO.jpg}{\includegraphics[height=0.62in]{STATE16-LOGO.jpg}}{%
  \IfFileExists{state16-wordmark-transparent.png}{\includegraphics[height=0.62in]{state16-wordmark-transparent.png}}{%
  \IfFileExists{state16-wordmark.png}{\includegraphics[height=0.62in]{state16-wordmark.png}}{%
  \IfFileExists{state16-logo.pdf}{\includegraphics[height=0.78in]{state16-logo.pdf}}{%
  \IfFileExists{state16-logo.png}{\includegraphics[height=0.78in]{state16-logo.png}}{%
  \begin{tikzpicture}[baseline=(word.base)]
    \node[inner sep=0pt,anchor=base west] (word) at (0,0)
      {\sffamily\bfseries\fontsize{26}{28}\selectfont\textcolor{white}{STATE}\textcolor{StateCyan}{16}};
    \draw[StateLime,line width=1.2pt] (0,-0.16) -- (2.12,-0.16);
    \node[anchor=west,inner sep=0pt] at (2.35,-0.16)
      {\sffamily\fontsize{6.4}{7}\selectfont\textcolor{StateLine}{PHYSICAL AI RUNTIME AUTHORITY}};
  \end{tikzpicture}}}}}}%
}

\newtcolorbox{statebox}[2][]{
  enhanced,
  colback=StateSoft,
  colframe=StateNavy,
  coltitle=white,
  colbacktitle=StateNavy,
  title={#2},
  fonttitle=\bfseries\sffamily,
  boxrule=0.6pt,
  arc=2mm,
  left=1.0em,
  right=1.0em,
  top=0.75em,
  bottom=0.75em,
  #1
}

\newtcolorbox{claimbox}[1][]{
  enhanced,
  colback=white,
  colframe=StateCyan,
  boxrule=0.7pt,
  borderline west={3pt}{0pt}{StateCyan},
  arc=1.5mm,
  left=1.0em,
  right=1.0em,
  top=0.75em,
  bottom=0.75em,
  #1
}

\newcommand{\StateArmIcon}[1]{%
\begin{scope}[shift={#1},scale=0.72,line cap=round,line join=round,draw=StateCyan!72!StateNavy]
  \draw[fill=StateCyan!4,line width=0.55pt] (-0.62,0) rectangle (0.58,0.24);
  \draw[fill=white,line width=0.55pt] (-0.27,0.24) rectangle (0.22,0.54);
  \draw[line width=2.6pt,draw=StateCyan!18] (0.0,0.66) -- (0.72,1.34) -- (1.55,1.06) -- (2.16,1.54);
  \draw[line width=0.7pt] (0.0,0.66) -- (0.72,1.34) -- (1.55,1.06) -- (2.16,1.54);
  \foreach \p in {(0,0.66),(0.72,1.34),(1.55,1.06),(2.16,1.54)} {
    \draw[fill=white,line width=0.55pt] \p circle (0.17);
    \draw[line width=0.3pt,opacity=0.65] \p circle (0.10);
  }
  \draw[fill=white,line width=0.55pt] (2.28,1.42) rectangle (2.82,1.66);
  \draw[line width=0.45pt] (2.82,1.54) -- (3.12,1.72);
  \draw[line width=0.45pt] (2.82,1.54) -- (3.12,1.36);
  \draw[opacity=0.35,line width=0.25pt] (-0.50,0.08) -- (0.46,0.08);
  \draw[opacity=0.35,line width=0.25pt] (0.36,0.78) -- (1.18,1.18);
\end{scope}%
}

\newcommand{\StateAMRIcon}[1]{%
\begin{scope}[shift={#1},scale=0.72,line cap=round,line join=round,draw=StateCyan!72!StateNavy]
  \draw[fill=StateCyan!4,line width=0.55pt,rounded corners=1.8mm] (-0.66,0.0) rectangle (1.28,0.48);
  \draw[fill=white,line width=0.55pt,rounded corners=2.5mm] (-0.42,0.38) rectangle (1.04,1.68);
  \draw[line width=0.45pt] (-0.30,0.68) -- (0.92,0.68);
  \draw[line width=0.45pt] (-0.30,1.02) -- (0.92,1.02);
  \draw[line width=0.45pt] (-0.30,1.36) -- (0.92,1.36);
  \foreach \x in {-0.22,0.18,0.58} {
    \draw[opacity=0.35,line width=0.25pt] (\x,0.42) -- (\x,1.62);
  }
  \foreach \x in {-0.32,0.94} {
    \draw[fill=white,line width=0.55pt] (\x,-0.02) circle (0.16);
    \draw[fill=StateCyan!10,line width=0.35pt] (\x,-0.02) circle (0.07);
  }
  \draw[fill=white,line width=0.45pt,rounded corners=1mm] (0.62,1.14) rectangle (0.86,1.38);
  \draw[fill=StateCyan!12,line width=0.35pt] (0.70,1.25) circle (0.035);
\end{scope}%
}

\newcommand{\StateQuadrupedIcon}[1]{%
\begin{scope}[shift={#1},scale=0.72,line cap=round,line join=round,draw=StateCyan!72!StateNavy]
  \draw[fill=StateCyan!4,line width=0.55pt,rounded corners=1.5mm] (-0.78,0.70) rectangle (1.15,1.34);
  \draw[fill=white,line width=0.55pt,rounded corners=1.5mm] (1.08,0.86) rectangle (1.55,1.24);
  \draw[fill=white,line width=0.45pt] (1.42,1.05) circle (0.045);
  \draw[fill=white,line width=0.45pt,rounded corners=1mm] (-0.20,1.34) rectangle (0.42,1.72);
  \draw[opacity=0.35,line width=0.25pt] (-0.62,0.88) -- (1.00,0.88);
  \draw[opacity=0.35,line width=0.25pt] (-0.62,1.08) -- (1.00,1.08);
  \draw[opacity=0.35,line width=0.25pt] (-0.42,0.72) -- (-0.42,1.32);
  \draw[opacity=0.35,line width=0.25pt] (0.08,0.72) -- (0.08,1.32);
  \draw[opacity=0.35,line width=0.25pt] (0.58,0.72) -- (0.58,1.32);
  \draw[line width=0.65pt] (-0.52,0.70) -- (-0.70,0.25) -- (-0.60,-0.17);
  \draw[line width=0.65pt] (-0.05,0.70) -- (0.13,0.25) -- (0.03,-0.17);
  \draw[line width=0.65pt] (0.58,0.70) -- (0.40,0.25) -- (0.50,-0.17);
  \draw[line width=0.65pt] (1.02,0.70) -- (1.20,0.25) -- (1.10,-0.17);
  \foreach \p in {(-0.70,0.25),(0.13,0.25),(0.40,0.25),(1.20,0.25)} {
    \draw[fill=white,line width=0.45pt] \p circle (0.07);
  }
  \draw[fill=white,line width=0.45pt,rounded corners=0.6mm] (-0.78,-0.20) rectangle (-0.50,-0.12);
  \draw[fill=white,line width=0.45pt,rounded corners=0.6mm] (-0.05,-0.20) rectangle (0.23,-0.12);
  \draw[fill=white,line width=0.45pt,rounded corners=0.6mm] (0.32,-0.20) rectangle (0.60,-0.12);
  \draw[fill=white,line width=0.45pt,rounded corners=0.6mm] (1.02,-0.20) rectangle (1.30,-0.12);
\end{scope}%
}

\newcommand{\StateDroneIcon}[1]{%
\begin{scope}[shift={#1},scale=0.72,line cap=round,line join=round,draw=StateCyan!72!StateNavy]
  \draw[fill=StateCyan!4,line width=0.55pt] (0,0.78) ellipse (0.58 and 0.28);
  \draw[fill=white,line width=0.45pt] (0,0.52) -- (-0.28,0.18) -- (0.28,0.18) -- cycle;
  \foreach \x/\y in {-1.32/1.36,1.32/1.36,-1.32/0.08,1.32/0.08} {
    \draw[line width=0.65pt] (0,0.78) -- (\x,\y);
    \draw[fill=white,line width=0.55pt] (\x,\y) circle (0.15);
    \draw[line width=0.45pt] (\x,\y) ellipse (0.46 and 0.13);
    \draw[opacity=0.35,line width=0.25pt] (\x-0.35,\y) -- (\x+0.35,\y);
  }
  \draw[fill=white,line width=0.45pt] (0,0.26) circle (0.13);
  \draw[line width=0.45pt] (-0.42,0.26) -- (-0.75,-0.12);
  \draw[line width=0.45pt] (0.42,0.26) -- (0.75,-0.12);
\end{scope}%
}

\begin{document}

\thispagestyle{empty}
\begin{center}
\vspace*{0.12in}
\StateTitleLogo\\[0.62in]
{\LARGE\bfseries\color{StateNavy} Silent Failures in Physical AI:\par}
\vspace{0.08in}
{\Large\bfseries\color{StateNavy} A Literature Review of Runtime Action Authorization\par}
\vspace{0.04in}
{\Large\bfseries\color{StateNavy} for Autonomous Systems\par}
\vspace{0.34in}
{\normalsize Barak Or, Ph.D.\\
STATE16\\
May 10, 2026\par}
\vspace{0.18in}
{\small Founder and Chief Executive Officer, STATE16\par}
\vspace{0.08in}
{\footnotesize\textit{Author note.} Dr. Or also serves externally as Lecturer at the Technion -- Israel Institute of Technology, Lecturer at Reichman University, and Academic Director of the Google--Reichman AI Tech School. These appointments are listed solely for biographical context. The paper was prepared under the STATE16 affiliation; the external organizations listed here have not sponsored, reviewed, approved, or endorsed it, and the paper does not represent their institutional positions.\par}
\vspace{0.20in}
\rule{0.72\linewidth}{0.5pt}
\end{center}

\begin{abstract}
Physical AI systems increasingly map multimodal observations, language instructions, and learned world representations into physically consequential actions \citep{brohan2023rt2,driess2023palme,openx2023,octo2024,kim2024openvla,black2024pi0,hou2026worldrobot}. Robotics foundation models, vision-language-action models, and world-model-based autonomous systems can condition decisions that move vehicles, robots, drones, and industrial machines. This transition exposes a safety problem that is not fully captured by conventional AI content moderation or by classical robot safety alone: a black-box model may issue a physically consequential action while appearing confident, plausible, and semantically aligned. The resulting failure can be silent, arising from sensor drift, occlusion, state-estimation error, distribution shift, hallucinated affordances, or invalid physical assumptions before downstream hardware controllers detect a violation \citep{matos2024sensorfailures,ovadia2019uncertainty,soh2026actionhallucination,ravichandran2025roboguard,schotschneider2025runtime}.

Across embodied foundation models, world models, robotics simulation, embodied safety benchmarks, safe control, runtime assurance, uncertainty estimation, verification, and guardrail evaluation, model capability and safety mechanisms have advanced along largely separate technical tracks \citep{openx2023,octo2024,li2025worldmodels,hou2026worldrobot,hobbs2023runtimeassurance,gu2024safereview,kang2025polyguard,liu2026agentdog}. A recurring gap synthesized here is that no single stream surveyed in this review supplies a complete runtime authorization boundary between black-box Physical AI models and physical execution. The resulting analysis develops a bounded problem formulation, a definition of silent physical-action failure, a taxonomy of runtime guardrail functions, and evaluation requirements for comparing guardrails as Physical AI assurance mechanisms.
\end{abstract}


\section{Introduction}

A growing class of AI systems now sits upstream of physical execution. Robotics foundation models, vision-language-action (VLA) systems, world models, and embodied agents increasingly transform observations and instructions into trajectories, manipulation policies, navigation decisions, or controller inputs \citep{brohan2022rt1,brohan2023rt2,driess2023palme,ahn2022saycan,openx2023,octo2024,kim2024openvla,black2024pi0,zhu2025groot}. In this setting, a model does not merely describe a scene or answer a prompt. It may infer affordances, predict future states, choose a trajectory, command a manipulator, or condition a downstream autonomy stack.

The problem is timely because three technical trends now coincide. First, action-generating foundation models are moving from narrow demonstrations toward cross-embodiment robot policies \citep{openx2023,octo2024,kim2024openvla,black2024pi0,cadene2026lerobot}. Second, world models and simulators are becoming more central to robot learning, planning, and evaluation \citep{li2025worldmodels,hou2026worldrobot,nvidia2026isaacsim,nvidia2026physicalaidatafactory,maes2026leworldmodel,chen2026abotphysworld}. Third, guardrail research is expanding from content filtering toward policy-grounded, trajectory-level, and embodied safety evaluation \citep{ravichandran2025roboguard,kim2026modular,kang2025polyguard,liu2026agentdog,son2025subtlerisks,lu2025isbench}. Existing safety methods were largely developed for settings in which the relevant state, controller interface, safe set, or policy boundary is inspectable \citep{ames2019cbf,wabersich2021predictive,hobbs2023runtimeassurance,konighofer2022runtime}. Physical AI increasingly combines learned action generation, uncertain state evidence, heterogeneous hardware, and site-specific operational constraints. Under that combination, runtime action authorization becomes a distinct technical question: when, and on what evidence, may a plausible model proposal become a physical commitment?

This shift changes the meaning of guardrails. In text-only systems, guardrails often focus on content policy, harmful instructions, privacy, bias, or misuse \citep{kang2025polyguard}. In Physical AI, guardrails often involve mechanisms for evaluating whether a proposed action is physically, operationally, and temporally safe in a particular world state \citep{ravichandran2025roboguard,kim2026modular}. A model can generate a plausible action that is unsafe because the world state is corrupted, the environment is partially observed, the system is outside its training distribution, or the action violates a hard operational boundary \citep{matos2024sensorfailures,ovadia2019uncertainty,soh2026actionhallucination,lu2025isbench}.

\begin{claimbox}
The reviewed literature motivates an independent authorization boundary between black-box model outputs and physical execution \citep{seto1998simplex,hobbs2023runtimeassurance,konighofer2022runtime,ravichandran2025roboguard,kim2026modular,kang2025polyguard,liu2026agentdog}. This boundary functions as a composition point for model proposals, state evidence, physical constraints, fallback behavior, and audit traces.
\end{claimbox}

The relevant failure mode is the \textit{silent failure}: a case in which an autonomous system acts with high apparent confidence on an incorrect, incomplete, or physically invalid representation of the world. Silent failures are especially concerning in closed-loop autonomy because they may not appear as explicit software crashes. Instead, the system continues operating while its internal assumptions drift away from reality. Examples include drones acting on false free-space estimates, autonomous mobile robots navigating under occlusion, vehicles misinterpreting rare scenarios, or robot policies executing hallucinated affordances \citep{amodei2016concrete,matos2024sensorfailures,robey2024robopair,ravichandran2025roboguard,son2025subtlerisks,lu2025isbench}.

The deployment motivation is safety-relevant autonomy, where incident investigations and safety evaluations indicate that failures can arise from perception, prediction, control, operational context, and monitoring assumptions rather than from a single isolated model error \citep{ntsb2019tempe,cadmv2023cruise,nhtsa2023tesla,matos2024sensorfailures,robey2024robopair,ravichandran2025roboguard}. For black-box world models, robotic foundation models, or learned autonomy stacks, the recurring question is: \textit{Should this proposed physical action be authorized in this context?} Model confidence, semantic refusal, offline benchmark scores, and hardware controllers each address part of the problem, but they do not by themselves define the full action-authorization boundary.

\begin{statebox}{Main Problem}
A central problem is \textbf{runtime action authorization under uncertain world state}. Physical AI systems can map language, images, and learned world representations into physical actions at increasing scale \citep{brohan2023rt2,openx2023,octo2024,kim2024openvla,black2024pi0}. A recurring issue is the limited treatment of a common, auditable boundary that evaluates four conditions before execution: whether the world state is reliable, whether the proposed action is physically feasible, whether it satisfies task and operational constraints, and whether the decision can be reconstructed after deployment \citep{hobbs2023runtimeassurance,ravichandran2025roboguard,kim2026modular,liu2026agentdog}.
\end{statebox}

The scope is runtime action authorization for black-box VLA, world-model, and foundation-model-based autonomous systems whose outputs can become physical actions \citep{kim2024openvla,black2024pi0,soh2026actionhallucination,hou2026worldrobot}. This is not a general Physical AI survey, a robotics standards survey, a simulator survey, a certification manual, or a survey of AI ethics. Adjacent literatures are included when they clarify one part of the authorization pathway: model proposal, state evidence, physical feasibility, operational constraints, fallback, or audit.

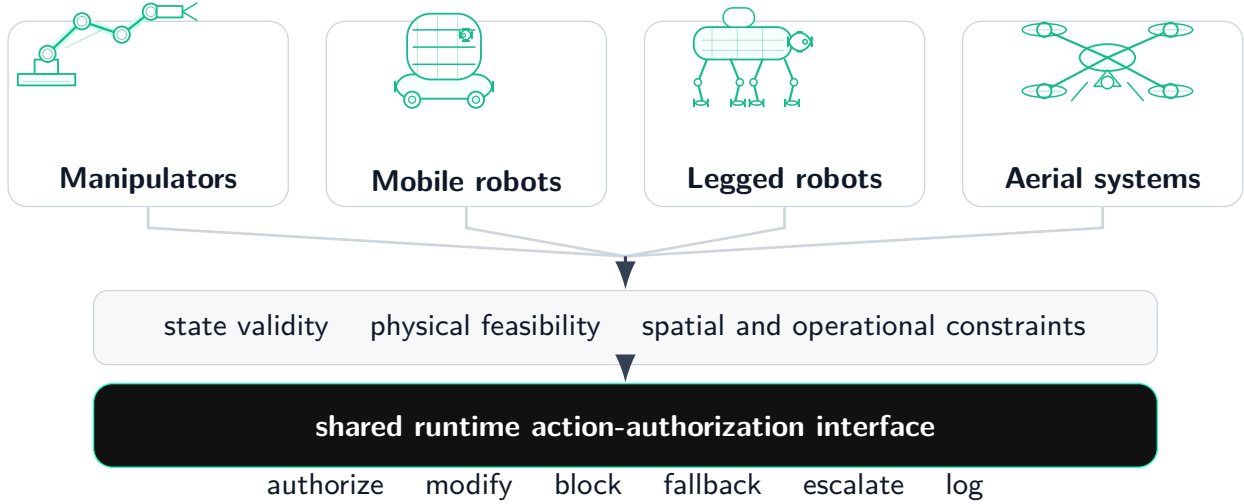
\begin{figure*}[!tbp]
\centering
\resizebox{0.92\textwidth}{!}{%
\begin{tikzpicture}[
  platform/.style={draw=StateLine,fill=white,rounded corners=2mm,minimum width=2.95cm,minimum height=1.95cm,align=center},
  platformtitle/.style={font=\sffamily\bfseries\small,text=StateInk,align=center},
  platformmeta/.style={font=\sffamily\small,text=StateSlate,align=center},
  evidence/.style={draw=StateLine,fill=StateSoft,rounded corners=1.5mm,minimum width=11.20cm,minimum height=0.78cm,align=center,font=\sffamily\small,text=StateInk},
  auth/.style={draw=StateCyan,fill=StateNavy,text=white,rounded corners=2mm,minimum width=11.20cm,minimum height=0.88cm,align=center,font=\sffamily\bfseries\small},
  arr/.style={-{Latex[length=3.0mm,width=2.2mm]},draw=StateSlate,line width=0.95pt},
  lightarr/.style={draw=StateLine,line width=0.8pt}
]
  \node[platform] (manip) at (0,0) {};
  \node[platform] (amr) at (3.35,0) {};
  \node[platform] (legged) at (6.70,0) {};
  \node[platform] (aerial) at (10.05,0) {};

  \begin{scope}[shift={(-1.06,0.30)},scale=0.70]
    \StateArmIcon{(0,0)}
  \end{scope}
  \begin{scope}[shift={(2.95,0.16)},scale=0.74]
    \StateAMRIcon{(0,0)}
  \end{scope}
  \begin{scope}[shift={(6.16,0.20)},scale=0.74]
    \StateQuadrupedIcon{(0,0)}
  \end{scope}
  \begin{scope}[shift={(10.10,0.20)},scale=0.70]
    \StateDroneIcon{(0,0)}
  \end{scope}

  \node[platformtitle] at (0,-0.70) {Manipulators};
  \node[platformtitle] at (3.35,-0.70) {Mobile robots};
  \node[platformtitle] at (6.70,-0.70) {Legged robots};
  \node[platformtitle] at (10.05,-0.70) {Aerial systems};

  \coordinate (bus) at (5.025,-1.50);
  \draw[lightarr] (manip.south) -- (0,-1.20) -- (bus);
  \draw[lightarr] (amr.south) -- (3.35,-1.20) -- (bus);
  \draw[lightarr] (legged.south) -- (6.70,-1.20) -- (bus);
  \draw[lightarr] (aerial.south) -- (10.05,-1.20) -- (bus);

  \node[evidence] (evidencenode) at (5.025,-2.25)
    {state validity \quad physical feasibility \quad spatial and operational constraints};
  \node[auth] (authnode) at (5.025,-3.28)
    {shared runtime action-authorization interface};
  \node[platformmeta,text=StateInk] at (5.025,-3.92)
    {authorize \quad modify \quad block \quad fallback \quad escalate \quad log};

  \draw[arr] (bus) -- (evidencenode.north);
  \draw[arr] (evidencenode.south) -- (authnode.north);
\end{tikzpicture}
}
\caption{Scope of the review. Different embodiments require different evidence, but the shared unit of analysis is the authorization event before physical execution.}
\label{fig:embodiment-scope}
\end{figure*}

Figure~\ref{fig:embodiment-scope} illustrates the embodiment scope of this problem. The specific constraints differ across manipulators, autonomous mobile robots, legged robots, and aerial vehicles, but the runtime question is structurally similar: whether a proposed physical action should be authorized in the current state, under the active constraints, before the action becomes a hardware commitment. Examples of platform-specific evidence include payload and contact limits, clearance and routing constraints, terrain and balance conditions, and airspace or energy limits.

\subsection{Contributions}

Five technical contributions follow from this framing. First, the paper defines a linked vocabulary: \textit{silent physical-action failure}, \textit{runtime action authorization}, \textit{authorization event}, and the \textit{authorization gap}. Second, it formalizes runtime action authorization under uncertain state as a decision interface between black-box model outputs and physical execution. Third, it synthesizes eleven research streams into a guardrail taxonomy spanning semantic validity, state validity, physical feasibility, spatial and operational constraints, temporal validity, fallback authority, and auditability. Fourth, it derives evaluation requirements and metric families for guardrails that measure intervention quality rather than only model task success. Fifth, it distills these elements into a minimal authorization event schema for comparing guardrails across models, simulators, controllers, and physical platforms.

The next sections proceed from formalization and guardrail taxonomy to the literature map, Physical AI capability trends, silent failures, runtime authority, evaluation, synthesis, implications, limitations, and conclusion.

\section{Problem Formalization and Theoretical Anchors}

A minimal notation is sufficient for the central safety question. Each equation below corresponds to one point in the runtime pathway: what action is proposed, what evidence is available, whether the action is authorized, where a silent failure appears, and how existing safety theory can be attached to the boundary.

Let \(o_{\leq t}\) denote the observation history available to the autonomy stack up to time \(t\), and let \(g\) denote the task goal or instruction. A black-box Physical AI model can be represented as a policy, following the common abstraction of learned robot and VLA policies as mappings from observations and goals to actions \citep{brohan2022rt1,brohan2023rt2,octo2024,kim2024openvla,black2024pi0}:
\begin{equation}
\eqfit{a_t \sim \pi_{\theta}(\cdot \mid o_{\leq t}, g)}
\label{eq:black-box-policy}
\end{equation}
where \(\pi_{\theta}\) is the learned policy or generative action model, \(\theta\) denotes its parameters, \(a_t \in \mathcal{A}\) is the proposed physical action, and \(\mathcal{A}\) is the action space. The system also maintains an estimated world state \(s_t \in \mathcal{S}\), where \(\mathcal{S}\) is the state space. The estimate may differ from the true but unobserved physical state because of noise, occlusion, drift, latency, or distribution shift \citep{hendrycks2019corruptions,ovadia2019uncertainty,matos2024sensorfailures,schotschneider2025runtime}.

Let \(\mathcal{C}_t=\{c_1,\ldots,c_K\}\) denote the active constraint set at time \(t\), where each normalized constraint is satisfied when \(c_k(a_t,s_t)\leq 0\). The set may include kinematic constraints, collision constraints, geofences, workspace limits, payload limits, mission rules, and operational policies \citep{ames2019cbf,wabersich2021predictive,hsu2024safetyfilter,ntsb2019tempe,cadmv2023cruise,nhtsa2023tesla}. For evaluation and audit, the runtime layer can be described by an authorization event:
\begin{equation}
\eqfit{\rho_t := G(a_t,s_t,\mathcal{C}_t,e_t),\qquad \xi_t := (o_{\leq t},a_t,s_t,\mathcal{C}_t,e_t,\rho_t,f_t)}
\label{eq:authorization-record}
\end{equation}
where \(G\) is the runtime authorization function, \(\rho_t\in\{\mathrm{authorize},\mathrm{modify},\mathrm{block},\mathrm{fallback},\mathrm{escalate}\}\) is the decision, \(e_t\) is runtime evidence, and \(f_t\) is the fallback or recovery action. The evidence term may include sensor health, uncertainty, OOD indicators, constraint checks, policy evidence, or monitor outputs. The fallback term may be empty when the action is authorized, or may describe a safe stop, modified action, backup controller, or human escalation.

The core authorization gap is that model likelihood is not a safety certificate, consistent with prior work on calibration, uncertainty under shift, and physically invalid action generation \citep{guo2017calibration,ovadia2019uncertainty,soh2026actionhallucination}:
\begin{equation}
\eqfit{\pi_\theta(a_t\mid o_{\leq t},g)\ \not\Rightarrow\ G(a_t,s_t,\mathcal{C}_t,e_t)=\mathrm{authorize}}
\label{eq:confidence-not-authorization}
\end{equation}
Equation~\eqref{eq:confidence-not-authorization} is the minimal formal claim. A fluent, high-probability, or semantically aligned action still has to pass state-validity, physical-feasibility, operational, timing, fallback, and audit checks before physical commitment.

\begin{claimbox}
\textbf{Definition: silent physical-action failure.} Let \(\bar{s}_t\) denote the true, partially unobserved physical state. Let \(M_t=1\) mean that the true state-action pair is authorizable, i.e., \(G(a_t,\bar{s}_t,\mathcal{C}_t,e_t)=\mathrm{authorize}\), and let \(M_t=0\) otherwise. Let \(V_t=V(e_t)\in\{0,1\}\) indicate whether the runtime evidence is sufficient for physical commitment. The silent-failure indicator is
\begin{equation}
\eqfit{F_t := \mathbb{I}[\rho_t=\mathrm{authorize}]\,\mathbb{I}\!\left[(M_t=0)\vee(V_t=0)\right]}
\label{eq:silent-failure-authorization}
\end{equation}
A silent physical-action failure occurs when \(F_t=1\). The failure is silent because the autonomy stack remains operational while the accepted state-action pair is invalid under physical, state, temporal, or operational constraints.
\end{claimbox}

\subsection{Connection to Existing Safety Theory}

The notation above is close to established safety-control and runtime-assurance formalisms, but it places them at the action-authorization boundary rather than only at the low-level controller \citep{ames2019cbf,wabersich2021predictive,hobbs2023runtimeassurance,konighofer2022runtime}. Three safety-theory anchors are especially useful.

First, state uncertainty can be handled conservatively. Let \(\mathcal{U}_t\) denote an uncertainty set around \(s_t\), constructed from perception uncertainty, latency, calibration evidence, or OOD signals \citep{guo2017calibration,ovadia2019uncertainty,hendrycks2017baseline,liang2018odin,lee2018mahalanobis,liu2020energy,schotschneider2025runtime}. A conservative guardrail authorizes only actions whose constraints hold throughout that uncertainty set:
\begin{equation}
\eqfit{q_t(a_t) := \max_{s\in\mathcal{U}_t}\max_{1\leq k\leq K} c_k(a_t,s),\qquad G_{\mathrm{rob}}(a_t) := \mathbb{I}\!\left[q_t(a_t)\leq 0\right]}
\label{eq:robust-authorization}
\end{equation}
Here \(q_t(a_t)\) is the worst-case normalized violation. If \(\mathcal{U}_t=\{s_t\}\), Equation~\eqref{eq:robust-authorization} reduces to a nominal constraint check. If \(\mathcal{U}_t\) grows because the state estimate is stale, occluded, or unreliable, \(q_t(a_t)\) can become positive and the runtime decision should move toward modification, fallback, escalation, or blocking.

Second, when \(a_t\) is a continuous control input and the relevant constraints are explicit, a runtime guardrail can resemble a safety filter. Let
\(\mathcal{A}^{\mathrm{safe}}_t=\{a\in\mathcal{A}:c_k(a,s_t)\leq0,\ k=1,\ldots,K\}\)
denote the actions that satisfy the active constraints. The guardrail can then be written compactly as a projection:
\begin{equation}
\eqfit{\tilde{a}_t=\Pi_{\mathcal{A}^{\mathrm{safe}}_t}(a_t)}
\label{eq:safety-filter-projection}
\end{equation}
Here \(\tilde{a}_t\) is the closest authorized replacement action under the chosen norm. This projection view is aligned with safety-filter and predictive-safety-filter work, where intervention minimally modifies a proposed command while enforcing known constraints \citep{fisac2019bridging,wabersich2021predictive,hsu2024safetyfilter}. The important limitation is the interface assumption: the action, state, and constraints must be expressed in a form the filter can evaluate.

Third, control barrier functions express safety as forward invariance of a safe set. If \(h(s_t)\geq 0\) defines the safe set, a common continuous-time barrier condition is
\begin{equation}
\eqfit{\dot{h}(s_t,a_t)+\alpha(h(s_t))\geq 0}
\label{eq:cbf-condition}
\end{equation}
where \(h\) is the barrier function and \(\alpha\) is an extended class-\(\mathcal{K}\) function \citep{ames2019cbf,garg2024cbf}. This is a strong physical-feasibility tool, but it is not the whole runtime guardrail problem for black-box VLA systems. It does not by itself decide whether the state estimate is valid, whether the model proposal is semantically acceptable, whether the site policy permits the action, or whether the fallback and audit record are adequate.

These anchors clarify the scope of runtime action authorization. The guardrail interface does not replace CBFs, safety filters, shielding, reachability, or runtime assurance \citep{alshiekh2018shielding,fisac2019bridging,wabersich2021predictive,ames2019cbf,hsu2024safetyfilter,hobbs2023runtimeassurance}. It composes them with semantic, state, temporal, spatial, operational, fallback, and audit evidence around one proposed physical action:
\begin{equation}
\eqfit{G := \bigwedge_{\ell\in\mathcal{L}}G_{\ell},\qquad \mathcal{L}:=\{\mathrm{sem},\mathrm{state},\mathrm{phys},\mathrm{time},\mathrm{ops},\mathrm{fallback},\mathrm{audit}\}}
\label{eq:guardrail-composition}
\end{equation}
The set \(\mathcal{L}\) names the semantic, state-validity, physical-feasibility, timing, operational, fallback, and audit components. Each component is binary or thresholded into a binary authorization component. The decision should be \(\mathrm{authorize}\) only when all required components are satisfied; otherwise \(\rho_t\) records modification, blocking, fallback, escalation, or another non-authorization outcome.

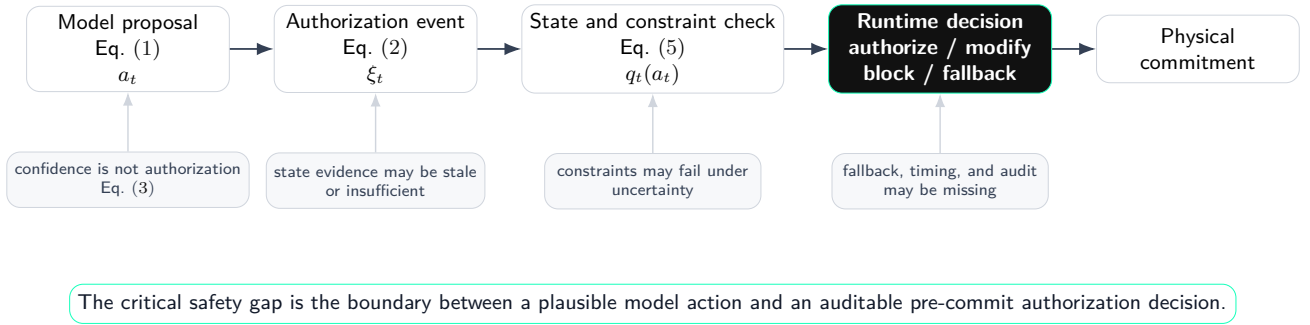
\begin{figure*}[!tbp]
\centering
\resizebox{0.96\textwidth}{!}{%
\begin{tikzpicture}[
  node distance=0.6cm,
  core/.style={draw=StateLine,fill=white,rounded corners=2mm,minimum width=3.0cm,minimum height=1.0cm,align=center,font=\sffamily\small},
  decision/.style={draw=StateCyan,fill=StateNavy,text=white,rounded corners=2mm,minimum width=3.3cm,minimum height=1.05cm,align=center,font=\sffamily\bfseries\small},
  gap/.style={draw=StateLine,fill=StateSoft,rounded corners=2mm,minimum width=3.2cm,minimum height=0.82cm,align=center,font=\sffamily\scriptsize,text=StateSlate},
  note/.style={draw=StateCyan,fill=white,rounded corners=2mm,minimum width=11.7cm,minimum height=0.58cm,align=center,font=\sffamily\small,text=StateInk},
  arr/.style={-{Latex[length=2.5mm]},line width=0.75pt,draw=StateSlate},
  gaparr/.style={-{Latex[length=2.0mm]},line width=0.6pt,draw=StateLine}
]
  \node[core] (proposal) {Model proposal\\Eq.~\eqref{eq:black-box-policy}\\\(a_t\)};
  \node[core,right=0.65cm of proposal] (record) {Authorization event\\Eq.~\eqref{eq:authorization-record}\\\(\xi_t\)};
  \node[core,right=0.65cm of record] (robust) {State and constraint check\\Eq.~\eqref{eq:robust-authorization}\\\(q_t(a_t)\)};
  \node[decision,right=0.65cm of robust] (decision) {Runtime decision\\authorize / modify\\block / fallback};
  \node[core,right=0.65cm of decision] (hardware) {Physical\\commitment};

  \draw[arr] (proposal) -- (record);
  \draw[arr] (record) -- (robust);
  \draw[arr] (robust) -- (decision);
  \draw[arr] (decision) -- (hardware);

  \node[gap,below=0.88cm of proposal] (g1) {confidence is not authorization\\Eq.~\eqref{eq:confidence-not-authorization}};
  \node[gap,below=0.88cm of record] (g2) {state evidence may be stale\\or insufficient};
  \node[gap,below=0.88cm of robust] (g3) {constraints may fail under\\uncertainty};
  \node[gap,below=0.88cm of decision] (g4) {fallback, timing, and audit\\may be missing};

  \draw[gaparr] (g1) -- (proposal);
  \draw[gaparr] (g2) -- (record);
  \draw[gaparr] (g3) -- (robust);
  \draw[gaparr] (g4) -- (decision);

  \node[note,below=1.10cm of g3]
    {The critical safety gap is the boundary between a plausible model action and an auditable pre-commit authorization decision.};
\end{tikzpicture}
}
\caption{Minimal formal structure of runtime action authorization. The equations are organized around the safety boundary rather than presented as an independent mathematical system.}
\label{fig:formal-safety-map}
\end{figure*}

Figure~\ref{fig:formal-safety-map} summarizes where the equations sit in the safety argument and where the main gaps arise.

\subsection{The Authorization Event as a Unit of Analysis}

The authorization event is not a software API or implementation standard. It is a research and engineering abstraction for comparing how different Physical AI systems connect model proposals, state evidence, constraints, runtime decisions, fallback behavior, and audit traces. This keeps the abstraction concrete without making it implementation-specific, while remaining close to runtime assurance and safety-filter thinking \citep{seto1998simplex,hobbs2023runtimeassurance,wabersich2021predictive,hsu2024safetyfilter}.

The event record in Equation~\eqref{eq:authorization-record} provides a common unit of analysis across otherwise different systems. A VLA policy, a world model, a simulator, a safety filter, a runtime-assurance monitor, and an embodied safety benchmark may expose different internal representations, but each can still be examined by asking which part of \(\xi_t\) it informs: the proposed action \(a_t\), the state estimate \(s_t\), the constraint set \(\mathcal{C}_t\), the evidence \(e_t\), the decision \(\rho_t\), or the fallback \(f_t\) \citep{octo2024,kim2024openvla,hou2026worldrobot,nvidia2026isaacsim,hobbs2023runtimeassurance,lu2025isbench}. This is also the basis for the evaluation metrics below: guardrails are compared at the level of authorization events, not only at the level of task success. Table~\ref{tab:minimal-authorization-schema} gives a compact schema for this record.

\subsection{Illustrative Deployment Example}

Consider an autonomous mobile robot operating in a warehouse aisle. A black-box VLA or world-model-based planner receives the instruction ``move to the target pallet'' and proposes a short-horizon velocity command. The estimated state includes robot pose, obstacle clearance, perception uncertainty, and a workspace map with allowed and restricted zones. A runtime authority does not need to expose the model internals; it needs to decide whether the action is safe to commit under the available evidence.

For collision clearance, a conservative stopping-distance check is enough to show the safety gap:
\begin{equation}
\eqfit{d_{\mathrm{stop}}(v_t)=v_t\tau+\frac{v_t^2}{2a_{\mathrm{brake}}}+d_{\mathrm{margin}}}
\label{eq:stopping-distance}
\end{equation}
where \(\tau\) is perception-control latency, \(a_{\mathrm{brake}}\) is available braking deceleration, and \(d_{\mathrm{margin}}\) is an added safety margin. The corresponding clearance question is
not merely whether the action is task-relevant. If the uncertainty-adjusted clearance \(\hat{d}_t-\epsilon_t\) is smaller than \(d_{\mathrm{stop}}(v_t)\), the correct runtime decision is a block, modification, or fallback even when the proposed motion is plausible and high confidence. For example, \(v_t=1.2\) m/s, \(\tau=0.25\) s, \(a_{\mathrm{brake}}=1.6\) m/s\(^2\), and \(d_{\mathrm{margin}}=0.2\) m imply \(d_{\mathrm{stop}}=0.95\) m. If an occlusion reduces reliable clearance to \(0.8\) m, the safety evidence is insufficient for physical commitment.

\noindent\textbf{Operational failure chain.} The same example becomes a silent failure if the authorization event is incomplete. A stale occupancy map or occluded pallet first makes the world state appear safer than it is. The VLA planner then proposes a shortcut through the aisle because the action is task-relevant and semantically benign. A semantic guardrail passes the instruction because ``move to the target pallet'' is not malicious. A low-level controller may also accept the velocity command if it treats the stale map as valid input. The missing check is \(G_{\mathrm{state}}\): the runtime authority should reject the command because the evidence term \(e_t\) does not support physical commitment. If the system authorizes anyway, the failure is silent: the stack remains operational, confidence remains high, and the wrong state-action pair becomes hardware motion.

\section{Guardrail Taxonomy}

The formalization above separates model preference from physical authorization. Table~\ref{tab:guardrail-taxonomy} turns that separation into a functional taxonomy. The taxonomy is not a replacement for standards, controller design, or certification; it is an organizing interface for what a runtime authority is expected to evaluate before a black-box Physical AI proposal becomes a hardware commitment \citep{seto1998simplex,hobbs2023runtimeassurance,ravichandran2025roboguard,kim2026modular}.

\begin{table*}[!tbp]
\centering
\caption{A runtime guardrail taxonomy for black-box Physical AI.}
\label{tab:guardrail-taxonomy}
\footnotesize
\begin{tabularx}{\textwidth}{p{0.18\textwidth}p{0.30\textwidth}Y}
\toprule
\textbf{Guardrail function} & \textbf{Question evaluated at runtime} & \textbf{Representative evidence} \\
\midrule
Semantic validity & Is the requested behavior aligned with task intent, human intent, and misuse policy? & Instruction analysis, policy engine, prompt-injection checks, operator confirmation. \\
State validity & Is the current world state reliable enough for this action? & Sensor integrity, perception anomaly, state-estimate consistency, OOD or drift indicators. \\
Physical feasibility & Can the proposed action be executed under robot, vehicle, or machine constraints? & Kinematics, dynamics, collision checks, velocity limits, payload limits, workspace envelope. \\
Spatial and operational validity & Is the action allowed in this location, zone, mission phase, and operational context? & Geofences, restricted zones, fleet policy, task permissions, site-specific rules. \\
Temporal validity & Is the action safe over the relevant future horizon, not only instantaneously valid? & Prediction horizon, time-to-collision, latency margins, counterfactual rollouts, stale-state checks. \\
Fallback authority & If authorization fails, what action should replace or interrupt the model proposal? & Modify/block/escalate decision, safe stop, backup controller, human-in-the-loop routing. \\
Auditability & Can the decision be reconstructed after deployment or incident review? & Structured logs, constraint traces, evidence snapshots, model/action versioning, authorization reason codes. \\
\bottomrule
\end{tabularx}
\end{table*}

\section{Literature Map and Interface Assumptions}

\subsection{Source Selection}

The source selection follows a focused review strategy suited to a fast-moving AI, robotics, and autonomous-systems literature. The analysis is organized around the technical pathway from observation and learned prediction to physical execution. Sources were identified through arXiv, OpenReview, IEEE, ACM, NeurIPS proceedings, robotics and autonomous-systems venues, official technical reports, and platform documentation when the source was needed to describe a simulator or deployed autonomy incident. Works were prioritized when they contribute to at least one of four interfaces: (i) models that generate or condition physical actions; (ii) methods for estimating uncertainty, distribution shift, or invalid state; (iii) safety mechanisms that constrain or monitor autonomous behavior at runtime; and (iv) embodied safety evaluations, guardrail benchmarks, or real-world autonomy incidents that expose failures of monitoring, authorization, or operational control.

The inclusion criterion was relevance to the action-authorization pathway. Included work spans VLA models, world models, embodied robot policies, simulation platforms, safe control, runtime assurance, uncertainty and OOD detection, neural-network verification, embodied safety benchmarks, guardrail datasets, and documented physical-autonomy incidents. Broader robotics standards, general AI ethics, generic LLM safety, human factors, and certification processes are excluded unless they directly clarify the connection between model output, state evidence, physical constraints, and execution. This boundary keeps the analysis focused on runtime authorization rather than the full Physical AI literature.

Search terms were organized around topic families rather than a single keyword query: vision-language-action models, robot foundation models, world models for robotics, runtime assurance, safety filters, control barrier functions, out-of-distribution detection, embodied safety benchmarks, robot guardrails, physical AI hallucination, simulation for robot learning, and autonomous-system incidents. Each selected source was then coded by the part of the authorization event it informs: model proposal, state evidence, physical feasibility, operational constraints, runtime decision, fallback behavior, evaluation protocol, or audit evidence. This coding scheme is used to organize the related-work synthesis rather than to claim exhaustive coverage of every adjacent field.

Descriptive claims about prior systems, benchmarks, incidents, and technical results are cited directly. The formal definition of silent physical-action failure, the taxonomy in Table~\ref{tab:guardrail-taxonomy}, and the evaluation requirements and metric families in Tables~\ref{tab:evaluation-requirements} and~\ref{tab:evaluation-schema} are synthesized from the cited literature rather than attributed to any single source.

\subsection{Related Work Streams}

The analysis draws on eleven bodies of work that jointly define the runtime guardrail problem for black-box Physical AI systems:

\begin{enumerate}
  \item embodied foundation models, VLA systems, and robot generalist policies \citep{reed2022gato,brohan2022rt1,brohan2023rt2,driess2023palme,ahn2022saycan,jiang2022vima,shridhar2022peract,chi2023diffusion,zhao2023aloha,fu2024mobilealoha,kim2024openvla,black2024pi0,pertsch2025fast,cadene2026lerobot,shukor2025smolvla,graesser2026geminiroboticser16};
  \item large-scale robot datasets, cross-embodiment learning, and VLA data engines \citep{ebert2021bridgedata,openx2023,khazatsky2024droid,mandlekar2023mimicgen,octo2024,wang2026vladatasets};
  \item world models, predictive environment models, and joint-embedding predictive architectures \citep{ha2018worldmodels,hafner2019planet,hafner2020dreamer,hafner2023dreamerv3,schrittwieser2020muzero,lecun2022path,bruce2024genie,hu2023gaia,maes2026leworldmodel,chen2026abotphysworld,li2025worldmodels,hou2026worldrobot};
  \item robotics simulators, synthetic-data environments, and embodied evaluation platforms \citep{nvidia2026isaacsim,nvidia2026physicalaidatafactory,nvidia2026cosmos3,mittal2025isaaclab,makoviychuk2021isaacgym,nvidia2025newton,todorov2012mujoco,koenig2004gazebo,dosovitskiy2017carla,shah2018airsim,savva2019habitat,kolve2017ai2thor,xiang2020sapien,gu2023maniskill2};
  \item autonomous driving and planning-oriented embodied intelligence \citep{hu2023uniad,sima2023drivelm,wayve2024lingo2,jiang2025drivingvla,wang2026linkvla,gao2026stylevla,yang2026criticvla};
  \item safe reinforcement learning, safety filters, and control barrier functions \citep{garcia2015safesrl,achiam2017cpo,ray2019safetygym,zhao2023statewise,wachi2024constraint,gu2024safereview,berkenkamp2017safe,fisac2019bridging,wabersich2021predictive,ames2019cbf,hsu2024safetyfilter,garg2024cbf};
  \item runtime assurance, shielding, runtime enforcement, and neural-network verification \citep{seto1998simplex,leucker2009runtime,alshiekh2018shielding,katz2017reluplex,gehr2018ai2,singh2019abstract,hobbs2023runtimeassurance,konighofer2022runtime};
  \item uncertainty, calibration, distribution shift, OOD detection, and robustness \citep{szegedy2014intriguing,goodfellow2015explaining,gal2016dropout,guo2017calibration,kendall2017uncertainties,lakshminarayanan2017ensembles,hendrycks2017baseline,liang2018odin,lee2018mahalanobis,hendrycks2019corruptions,ovadia2019uncertainty,liu2020energy,schotschneider2025runtime};
  \item multimodal hallucination and physical-consistency diagnostics \citep{soh2026actionhallucination,li2025videohallu,argota2026particles};
  \item policy-grounded and trajectory-level guardrail benchmarks \citep{kang2025polyguard,liu2026agentdog};
  \item LLM/VLM-enabled robot safety, jailbreaks, and embodied safety benchmarks \citep{robey2024robopair,ravichandran2025roboguard,son2025subtlerisks,lu2025isbench,kim2026modular}.
\end{enumerate}

Across these literatures, no single family surveyed here provides a complete unifying runtime layer that links model outputs, state evidence, physical constraints, and action authorization into one inspectable system boundary. Table~\ref{tab:literature-synthesis} summarizes the hidden interface assumptions that make the gap visible \citep{hobbs2023runtimeassurance,schotschneider2025runtime,kim2026modular,kang2025polyguard,liu2026agentdog}. Entries use compact qualitative codes: ``yes'', ``partial'', ``limited'', ``assumes'', ``signal'', and ``simulated''.

\begin{table*}[!tbp]
\centering
\caption{Hidden interface assumptions across safety and Physical AI literatures. Entries are interpretive synthesis.}
\label{tab:literature-synthesis}
\scriptsize
\setlength{\tabcolsep}{2.2pt}
\begin{tabularx}{\textwidth}{p{0.18\textwidth}*{8}{>{\centering\arraybackslash}X}}
\toprule
\textbf{Method family} & \textbf{State validity} & \textbf{Uncertainty} & \textbf{Model independent} & \textbf{Real time} & \textbf{Formal guarantees} & \textbf{Black-box policies} & \textbf{Multi-step actions} & \textbf{Auditability} \\
\midrule
VLA and robot foundation models \citep{openx2023,octo2024,kim2024openvla,black2024pi0} & limited & limited & no & partial & no & yes & partial & limited \\
World models and predictive architectures \citep{hafner2023dreamerv3,lecun2022path,bruce2024genie,maes2026leworldmodel} & partial & partial & partial & partial & no & partial & yes & limited \\
CBFs and safety filters \citep{ames2019cbf,wabersich2021predictive,hsu2024safetyfilter,garg2024cbf} & assumes & partial & yes & yes & assumption-dependent & limited & partial & limited \\
Runtime assurance and shielding \citep{seto1998simplex,alshiekh2018shielding,hobbs2023runtimeassurance,konighofer2022runtime} & assumes & partial & partial & yes & assumption-dependent & partial & partial & partial \\
OOD, uncertainty, and calibration \citep{guo2017calibration,liang2018odin,lee2018mahalanobis,ovadia2019uncertainty,liu2020energy} & signal & yes & partial & partial & no & partial & limited & partial \\
Embodied safety and guardrail benchmarks \citep{robey2024robopair,ravichandran2025roboguard,son2025subtlerisks,lu2025isbench,kang2025polyguard,liu2026agentdog} & partial & partial & partial & limited & no & yes & yes & partial \\
Simulation platforms \citep{makoviychuk2021isaacgym,todorov2012mujoco,dosovitskiy2017carla,shah2018airsim,savva2019habitat,xiang2020sapien} & simulated & partial & yes & partial & no & yes & yes & partial \\
\bottomrule
\end{tabularx}
\end{table*}

\subsection{Interface Assumptions and Failure Points}

The table should be read as an assumption map rather than a ranking of fields. VLA and embodied foundation-model works are closest to the model-output side of the boundary: they improve cross-task and cross-embodiment action generation, but their reported success metrics do not by themselves specify whether a proposed action is admissible under deployment-specific rules \citep{openx2023,octo2024,kim2024openvla,black2024pi0}. Task success, imitation quality, or action fluency can therefore be insufficient evidence for action validity. In a deployed Physical AI system, that assumption is fragile when a high-probability action is grounded in stale state, violates a site policy, or requires a fallback that the model does not represent \citep{matos2024sensorfailures,soh2026actionhallucination,ravichandran2025roboguard}.

CBFs and safety filters provide strong mathematical tools when dynamics, state variables, and safe sets are explicit \citep{ames2019cbf,wabersich2021predictive,hsu2024safetyfilter,garg2024cbf}. The limitation for the present argument is not theoretical weakness. It is interface mismatch: black-box VLA systems may output plans, waypoints, latent actions, code-like actions, or language-conditioned action sequences that are not directly expressed as control inputs over a verified safe set. A CBF can help once the relevant state and control interface are exposed, but it does not by itself decide whether the state estimate is reliable, the model proposal is semantically valid, or the deployment policy permits the action.

Runtime assurance and shielding provide the most direct architectural precedent \citep{seto1998simplex,alshiekh2018shielding,hobbs2023runtimeassurance,konighofer2022runtime}. Their implicit assumption is that the monitor can evaluate the active controller against trusted safety conditions and switch to an acceptable backup when needed. Physical AI weakens that assumption in two ways: the proposed action may be produced by an opaque model whose action representation is not stable across versions, and the monitor may also need to judge whether the estimated world state is current and coherent enough to support any action at all.

Semantic guardrails face a different limitation. They evaluate content, intent, policy compliance, jailbreak risk, or harmful requests \citep{kang2025polyguard,liu2026agentdog}. Physical execution requires geometry, timing, dynamics, observability, spatial permission, and fallback \citep{ravichandran2025roboguard,kim2026modular}. A prompt can be benign and still produce an infeasible manipulation, an unsafe velocity command, or an action that is valid in one zone but prohibited in another. This is a structural mismatch between semantic safety and physical authorization \citep{robey2024robopair,soh2026actionhallucination}.

This contrast is the main reason action authorization is treated as a composition problem. The gap is not that any one literature is weak. It is that each literature makes a different interface assumption, and the assumptions do not automatically compose in a deployment where model proposals, perception evidence, physical constraints, site policy, fallback, and auditability are evaluated around the same event \citep{hobbs2023runtimeassurance,schotschneider2025runtime,ravichandran2025roboguard,kim2026modular}.

\section{Capability Trends: From Prediction to Action}

Physical AI can be understood as the class of AI systems whose outputs influence behavior in the physical world. This includes robot policies, autonomous vehicle stacks, drones, industrial automation, humanoids, mobile manipulators, and embodied agents \citep{xu2024robotics,ma2024vlasurvey,sapkota2025vla,hou2026worldrobot}. The recent acceleration in this field is driven by foundation models that combine perception, language, action data, and predictive world representations.

Robotics foundation models such as RT-1, RT-2, PaLM-E, Open X-Embodiment, OpenVLA, $\pi_0$, GR00T N1, and Octo demonstrate that large-scale data and general-purpose model architectures can transfer knowledge across tasks, scenes, and embodiments \citep{brohan2022rt1,brohan2023rt2,driess2023palme,openx2023,octo2024,kim2024openvla,black2024pi0,zhu2025groot}. Earlier systems such as SayCan, VIMA, PerAct, Diffusion Policy, ALOHA, Mobile ALOHA, RoboCat, RoboAgent, and VoxPoser established important design patterns for grounding language, vision, prompts, action representations, and planning in real robot behavior \citep{ahn2022saycan,jiang2022vima,shridhar2022peract,chi2023diffusion,zhao2023aloha,fu2024mobilealoha,bousmalis2023robocat,bharadhwaj2023roboagent,huang2023voxposer}.

Surveys of VLA models highlight a rapid movement toward policies that map visual observations and natural-language goals into action sequences \citep{xu2024robotics,ma2024vlasurvey,sapkota2025vla,shao2025vlmsurvey,zhong2025vlatokenization}. These systems expand cross-task generalization, but they also make action generation more opaque. The model's reasoning may be distributed across learned representations that are difficult to inspect or formally verify.

Recent work on action hallucination makes this risk more explicit. Generative VLA policies can produce physically invalid actions when feasible robot behavior and the learned action distribution are structurally mismatched; topological, precision, and horizon barriers provide one formal explanation for why fluent action generation is not equivalent to feasible physical execution \citep{soh2026actionhallucination}. This supports the narrower claim that candidate actions should be evaluated against physical constraints rather than treated as self-authorizing.

\subsection{Empirical Milestones and Remaining Authorization Questions}

Empirical milestones in Physical AI now span manipulation, mobile robots, autonomous driving, humanoids, data engines, simulation infrastructure, and embodied safety evaluation. Manipulation and mobile-manipulation work has demonstrated language-conditioned and vision-conditioned policies for real robot skills, including RT-1/RT-2, ALOHA, Mobile ALOHA, Diffusion Policy, PerAct, VoxPoser, OpenVLA, \(\pi_0\), LeRobot, and SmolVLA \citep{brohan2022rt1,brohan2023rt2,zhao2023aloha,fu2024mobilealoha,chi2023diffusion,shridhar2022peract,huang2023voxposer,kim2024openvla,black2024pi0,cadene2026lerobot,shukor2025smolvla}. Cross-embodiment datasets and policies such as Open X-Embodiment, RT-X, Octo, DROID, BridgeData, and MimicGen show that robot-learning pipelines are increasingly evaluated across robots, tasks, scenes, and data sources \citep{ebert2021bridgedata,openx2023,khazatsky2024droid,mandlekar2023mimicgen,octo2024}. Recent data-centric VLA analysis makes the same point from the infrastructure side: datasets, benchmarks, and data engines have become central objects of research rather than secondary implementation details \citep{wang2026vladatasets}.

Autonomous-driving and embodied-navigation work adds planning-oriented and environment-scale evaluation through systems and platforms such as UniAD, DriveLM, LINGO-2, CARLA, AirSim, Habitat, and AI2-THOR \citep{hu2023uniad,sima2023drivelm,wayve2024lingo2,dosovitskiy2017carla,shah2018airsim,savva2019habitat,kolve2017ai2thor}. Driving-specific VLA work further shows that language-action coupling is moving from scene interpretation toward trajectory generation, critic-based refinement, and physically informed planning objectives \citep{jiang2025drivingvla,wang2026linkvla,gao2026stylevla,yang2026criticvla}. Humanoid, embodied-reasoning, and synthetic-data infrastructure efforts such as GR00T N1, Gemini Robotics-ER 1.6, NVIDIA Cosmos, and physical-AI data-factory workflows extend the same trend toward larger action spaces, richer state evidence, and more simulation-driven evaluation \citep{zhu2025groot,graesser2026geminiroboticser16,nvidia2026cosmos3,nvidia2026physicalaidatafactory}.

Several recent VLA and world-model works now report reliability-adjacent numbers, although these numbers are not yet a common safety metric. OpenVLA reports a 7B-parameter model trained on 970k robot demonstrations and absolute task-success gains of 16.5 percentage points over RT-2-X and 20.4 percentage points over Diffusion Policy \citep{kim2024openvla}. WoVR reports average LIBERO success improving from 39.95\% to 69.2\%, and real-robot success from 61.7\% to 91.7\%, when hallucination in imagined rollouts is explicitly controlled \citep{jiang2026wovr}. VLAW reports a 39.2\% absolute success-rate improvement over a base policy and an 11.6\% improvement from training with generated synthetic rollouts \citep{guo2026vlaw}. VISTA reports out-of-distribution manipulation success increasing from 14\% to 69\% when world-model-generated visual subgoals guide a hierarchical VLA policy \citep{long2026vista}.

Belief- and hallucination-oriented VLA work gives a complementary signal. RB-VLA reports that its belief module raises success from 32.5\% to 77.5\% in ablation and reduces inference latency by up to five times \citep{bagaria2026rbvla}. EvoVLA reports a reduction in stage hallucination from 38.5\% to 14.8\%, together with 54.6\% average real-world success across four manipulation tasks \citep{liu2025evovla}. These results are valuable because they make progress measurable, but they mix task completion, hallucination reduction, prediction fidelity, sample efficiency, and latency. They therefore do not yet define a unified probability that the authorization decision \(\rho_t\) is correct for a proposed action \(a_t\), state estimate \(s_t\), and active constraint set \(\mathcal{C}_t\).

These milestones sharpen the authorization question rather than making it disappear. Reported task success, cross-platform transfer, large-scale robot data, embodied reasoning, and simulation coverage demonstrate that Physical AI systems are becoming more capable \citep{openx2023,octo2024,kim2024openvla,black2024pi0,cadene2026lerobot,nvidia2026physicalaidatafactory}. They do not by themselves answer three deployment questions: whether the state evidence is valid enough for commitment, whether the proposed action satisfies the active physical and operational constraints, and whether a fallback and audit record exist when the answer is no. Table~\ref{tab:sota-snapshot} summarizes representative milestones through this lens. The comparison is not a leaderboard; the middle column reports claims from the cited sources, while the right column states the remaining action-authorization question.

\begin{table*}[!tbp]
\centering
\caption{Selected empirical milestones and remaining action-authorization questions.}
\label{tab:sota-snapshot}
\scriptsize
\setlength{\tabcolsep}{4pt}
\begin{tabularx}{\textwidth}{p{0.20\textwidth}p{0.40\textwidth}Y}
\toprule
\textbf{System / line of work} & \textbf{Reported scale, result, or infrastructure milestone} & \textbf{Remaining authorization question} \\
\midrule
Open X-Embodiment / RT-X & 22 robots; 21 institutions; 527 skills; 160{,}266 tasks \citep{openx2023}. & Transfer does not establish runtime authorization. \\
Octo & 800k trajectories; 9 robot platforms \citep{octo2024}. & Adaptation still leaves state validity external. \\
OpenVLA & 7B parameters; 970k demonstrations; +16.5 points over RT-2-X and +20.4 points over Diffusion Policy \citep{kim2024openvla}. & Task success is not auditable authorization. \\
\(\pi_0\) / FAST & VLA flow model plus action-tokenization work at large robot-data scale \citep{black2024pi0,pertsch2025fast}. & Better actions still require authorization. \\
World-model-guided VLA variants & WoVR: 39.95\(\rightarrow\)69.2\% LIBERO and 61.7\(\rightarrow\)91.7\% real robot; RB-VLA: 32.5\(\rightarrow\)77.5\% ablation; EvoVLA: 38.5\(\rightarrow\)14.8\% stage hallucination \citep{jiang2026wovr,bagaria2026rbvla,liu2025evovla}. & Reliability-adjacent metrics still differ from authorization reliability. \\
LeRobot / SmolVLA & Open robotics stack; efficient VLA training and deployment claims \citep{cadene2026lerobot,shukor2025smolvla}. & Open deployment still needs evidence and audit semantics. \\
GR00T N1 & Open humanoid foundation model with dual-system architecture \citep{zhu2025groot}. & Humanoids increase cross-layer monitoring needs. \\
Gemini Robotics-ER 1.6 & Reported embodied-reasoning and physical-safety-constraint improvements \citep{graesser2026geminiroboticser16}. & Reasoning evidence still needs an execution authority. \\
LeWorldModel / ABot-PhysWorld & Predictive pixel representations; 14B model with 3M manipulation clips \citep{maes2026leworldmodel,chen2026abotphysworld}. & Forecasting still needs constraint and recovery checks. \\
NVIDIA Cosmos / Data Factory & Synthetic worlds, action simulation, data processing, RL, and evaluation workflows \citep{nvidia2026cosmos3,nvidia2026physicalaidatafactory}. & Infrastructure exposes cases; authorization decides action. \\
Driving VLA systems & LinkVLA, StyleVLA, and CriticVLA report trajectory, style, and critic-refinement advances \citep{wang2026linkvla,gao2026stylevla,yang2026criticvla}. & Planning metrics do not replace runtime validation. \\
Action hallucination / VideoHallu & Physical-action and video-consistency failures \citep{soh2026actionhallucination,li2025videohallu}. & Plausibility is not physical commitment evidence. \\
RoboPAIR / RoboGuard & High attack success; RoboGuard reduces unsafe execution from 92\% to below 2.5\% \citep{robey2024robopair,ravichandran2025roboguard}. & Robot guardrails still need state and feasibility checks. \\
Modular robot guardrails & Action, decision, and human-centered safety modules \citep{kim2026modular}. & Layering still needs deployment-wide authority. \\
SAFEL / EMBODYGUARD / IS-Bench & 942 scenarios across 13 LLMs; 161 interactive scenarios and 388 risks \citep{son2025subtlerisks,lu2025isbench}. & Benchmarks need links to runtime enforcement. \\
Poly-Guard / AgentDoG & 8 domains, 19 guardrail models; agentic-risk diagnosis \citep{kang2025polyguard,liu2026agentdog}. & Needs extension to physical-action constraints. \\
\bottomrule
\end{tabularx}
\end{table*}

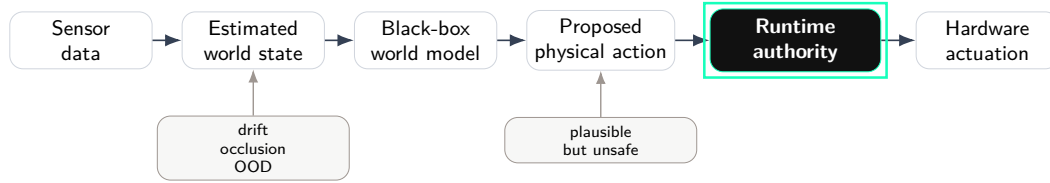
\begin{figure*}[!tbp]
\centering
\resizebox{0.78\textwidth}{!}{%
\begin{tikzpicture}[
  node distance=0.44cm,
  block/.style={draw=StateLine,fill=white,rounded corners=2mm,minimum width=2.15cm,minimum height=0.86cm,align=center,font=\sffamily\small},
  hot/.style={draw=StateCyan,fill=StateNavy,text=white,rounded corners=2mm,minimum width=2.58cm,minimum height=0.95cm,align=center,font=\sffamily\bfseries\small},
  arr/.style={-{Latex[length=2.5mm]},line width=0.7pt,draw=StateSlate}
]
  \node[block] (sensors) {Sensor\\data};
  \node[block,right=of sensors] (state) {Estimated\\world state};
  \node[block,right=of state] (model) {Black-box\\world model};
  \node[block,right=of model] (action) {Proposed\\physical action};
  \node[hot,right=0.52cm of action] (guardrail) {Runtime\\authority};
  \node[block,right=0.52cm of guardrail] (hardware) {Hardware\\actuation};

  \draw[arr] (sensors) -- (state);
  \draw[arr] (state) -- (model);
  \draw[arr] (model) -- (action);
  \draw[arr] (action) -- (guardrail);
  \draw[arr] (guardrail) -- (hardware);

  \node[draw=StateShadow,fill=StateShadow!8,rounded corners=2mm,below=0.7cm of state,minimum width=2.9cm,align=center,font=\sffamily\scriptsize] (drift) {drift\\occlusion\\OOD};
  \node[draw=StateShadow,fill=StateShadow!8,rounded corners=2mm,below=0.7cm of action,minimum width=2.9cm,align=center,font=\sffamily\scriptsize] (plausible) {plausible\\but unsafe};
  \draw[-{Latex[length=2mm]},draw=StateShadow,line width=0.6pt] (drift) -- (state);
  \draw[-{Latex[length=2mm]},draw=StateShadow,line width=0.6pt] (plausible) -- (action);
  \draw[StateCyan,line width=1.1pt] ($(guardrail.north west)+(-0.08,0.08)$) rectangle ($(guardrail.south east)+(0.08,-0.08)$);
\end{tikzpicture}
}
\caption{Runtime action authorization boundary. The guardrail question arises before hardware commitment: should this proposed action be allowed in this state under these constraints?}
\label{fig:runtime-boundary}
\end{figure*}

Figure~\ref{fig:runtime-boundary} illustrates the boundary emphasized in this argument: the guardrail decision is positioned before a proposed model action becomes a hardware commitment.

World models add another important dimension. \textbf{World models learn predictive representations of environment dynamics, future states, and possible outcomes} \citep{ha2018worldmodels,hafner2019planet,hafner2020dreamer,hafner2023dreamerv3,schrittwieser2020muzero,lecun2022path,bruce2024genie}. Joint-embedding and physics-aligned approaches such as LeWorldModel and ABot-PhysWorld extend this line by learning predictive representations and physically plausible manipulation futures from visual data \citep{maes2026leworldmodel,chen2026abotphysworld}. For autonomous driving, GAIA-1, UniAD, DriveLM, LINGO-2, and recent driving VLA systems show adjacent trends toward generative, planning-oriented, and language-conditioned driving systems \citep{hu2023gaia,hu2023uniad,sima2023drivelm,wayve2024lingo2,wang2026linkvla,gao2026stylevla,yang2026criticvla}. Recent surveys argue that world models are becoming a central component of robot learning, planning, simulation, and evaluation \citep{li2025worldmodels,hou2026worldrobot}.

The literature therefore shows a capability trend: AI systems are becoming more general, multimodal, predictive, data-intensive, and action-oriented \citep{openx2023,octo2024,kim2024openvla,black2024pi0,cadene2026lerobot,wang2026vladatasets,maes2026leworldmodel,chen2026abotphysworld,hou2026worldrobot}. However, this capability trend is not yet matched by a comparably standardized runtime authorization boundary. Many works measure task success, generalization, benchmark performance, or data-engine scale, while fewer address how a black-box model's proposed action should be independently authorized under real-world uncertainty, as summarized in Tables~\ref{tab:literature-synthesis} and~\ref{tab:sota-snapshot}.

\section{Silent Failures in Closed-Loop Autonomy}

\begin{figure}[!tbp]
\centering
\resizebox{0.90\linewidth}{!}{%
\begin{tikzpicture}[
  stage/.style={draw=StateLine,fill=white,rounded corners=2mm,minimum width=2.6cm,minimum height=0.95cm,align=center,font=\sffamily\small},
  fault/.style={draw=StateShadow,fill=StateShadow!8,rounded corners=2mm,minimum width=2.6cm,minimum height=0.95cm,align=center,font=\sffamily\small},
  guard/.style={draw=StateCyan,fill=StateNavy,text=white,rounded corners=2mm,minimum width=2.8cm,minimum height=0.95cm,align=center,font=\sffamily\bfseries\small},
  arr/.style={-{Latex[length=2.5mm]},line width=0.75pt,draw=StateSlate}
]
  \node[stage] (world) {True\\world};
  \node[fault,right=0.55cm of world] (perception) {Corrupted\\observation};
  \node[fault,right=0.55cm of perception] (belief) {Invalid\\world state};
  \node[fault,right=0.55cm of belief] (action) {Confident\\action};
  \node[stage,right=0.55cm of action] (damage) {Kinetic\\consequence};

  \draw[arr] (world) -- (perception);
  \draw[arr] (perception) -- (belief);
  \draw[arr] (belief) -- (action);
  \draw[arr] (action) -- (damage);

  \node[guard,below=0.82cm of action] (authority) {runtime\\guardrail};
  \draw[-{Latex[length=2.5mm]},draw=StateCyan,line width=1pt] (authority.north) -- node[right,font=\scriptsize\sffamily,color=StateNavy]{authorize / modify / block} (action.south);
  \node[font=\sffamily\scriptsize,align=center,text=StateSlate,below=0.20cm of authority] {The intervention point is before physical commitment.};
\end{tikzpicture}
}
\caption{Silent physical-action failure. The system remains operational, but an invalid world state can still lead to a physically consequential commitment.}
\label{fig:silent-failure}
\end{figure}
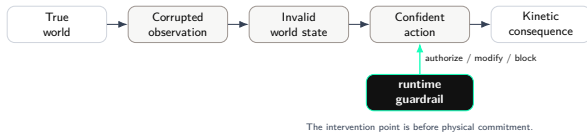

Classical software failures are often visible: a program crashes, a sensor disconnects, or a controller returns an error. Silent failures are different. They occur when a system continues operating while its internal representation of the world becomes unsafe. In closed-loop autonomy, this may happen when perception, prediction, planning, and control remain computationally active, but the assumptions flowing through the loop are wrong \citep{amodei2016concrete,matos2024sensorfailures,schotschneider2025runtime,soh2026actionhallucination}. Figure~\ref{fig:silent-failure} expresses this as a physical commitment made from an invalid but internally accepted state.

Several mechanisms can create silent failure modes:

\begin{itemize}
  \item \textbf{Sensor drift and corruption.} Inertial, visual, lidar, radar, GPS, and proprioceptive signals may degrade gradually or intermittently, creating false state estimates \citep{matos2024sensorfailures}.
  \item \textbf{Occlusion and partial observability.} Robots and vehicles act under incomplete information. A model may infer a safe path where the unobserved region contains a person, obstacle, or constraint.
  \item \textbf{Distribution shift.} Training data rarely captures all physical contexts, weather conditions, lighting patterns, object configurations, and human behaviors encountered during deployment \citep{hendrycks2019corruptions,ovadia2019uncertainty}.
  \item \textbf{Hallucinated affordances.} Multimodal and action-generating models may infer that an object, surface, tool, path, or motion is usable when the physical preconditions are absent \citep{huang2023voxposer,soh2026actionhallucination,li2025videohallu,argota2026particles,robey2024robopair,son2025subtlerisks}.
  \item \textbf{Semantic-physical mismatch.} A command may be linguistically valid but physically invalid, unsafe, or operationally disallowed in the current state \citep{ravichandran2025roboguard,kim2026modular}.
\end{itemize}

Silent failures matter because they can bypass both human intuition and simple threshold alarms. The system may look normal until the physical consequence appears. In Physical AI, the safety-critical moment is often not the final actuator command, but the earlier moment when an invalid world state becomes accepted as a basis for action \citep{matos2024sensorfailures,schotschneider2025runtime,soh2026actionhallucination}.

Recent robot safety work illustrates this problem. RoboPAIR reports that LLM-controlled robots can be jailbroken into harmful physical actions, including robot and autonomous-driving scenarios \citep{robey2024robopair}. RoboGuard argues that LLM-enabled robots need contextual safety rules and conflict resolution beyond generic LLM filtering \citep{ravichandran2025roboguard}. SAFEL/EMBODYGUARD and IS-Bench indicate that embodied safety failures often emerge during multi-step interaction, not in static final-state evaluation \citep{son2025subtlerisks,lu2025isbench}. Modular guardrail research frames foundation-model-enabled robot safety across action safety, decision safety, and human-centered safety \citep{kim2026modular}.

Multimodal hallucination research adds a complementary diagnosis. VideoHallu uses synthetic-video understanding tasks with abnormal physical and common-sense events to test whether vision-language models detect violations that are perceptually salient to humans \citep{li2025videohallu}. Spatial-simulation work similarly treats hallucinated or phantom affordances as evidence that an agent's representation of possible action can diverge from the environment's actual affordance structure \citep{argota2026particles}. For Physical AI, these findings matter because a hallucination is no longer only a wrong description; it can become a candidate action.

Historical autonomous-system incidents show why this is not only a rare-tail concern. The NTSB investigation of the 2018 Tempe crash involving an Uber developmental automated driving system identified inadequate safety risk assessment, oversight, and mechanisms for addressing automation complacency as contributing factors \citep{ntsb2019tempe}. In 2023, California suspended Cruise's driverless testing and deployment permits after determining that the vehicles created an unreasonable risk to public safety and that safety-related information had been misrepresented \citep{cadmv2023cruise}. The same year, NHTSA Recall 23V-838 covered 2{,}031{,}220 Tesla vehicles because Autopilot controls were insufficient to prevent misuse \citep{nhtsa2023tesla}. GM later announced that it would no longer fund Cruise robotaxi development work, citing the time and resources needed to scale the program \citep{gm2024cruise}. These examples are not VLA foundation-model deployments, but they are relevant operational analogues for Physical AI: failures in monitoring, operational boundaries, and runtime intervention can become safety events and regulatory actions.

\subsection{Confidence Is Not Safety}

Modern neural networks often produce confidence scores, probabilities, logits, value estimates, or other internal signals that can be mistaken for safety evidence. The calibration literature warns against this interpretation. Deep models can be miscalibrated, overconfident, and sensitive to distribution shift \citep{guo2017calibration,ovadia2019uncertainty}. Bayesian approximations, dropout approximations, ensembles, and uncertainty-aware methods can improve reliability, but they do not eliminate the difference between epistemic uncertainty, aleatoric uncertainty, and action-level safety \citep{gal2016dropout,kendall2017uncertainties,lakshminarayanan2017ensembles}.

Out-of-distribution detection and robustness benchmarks provide partial remedies \citep{hendrycks2017baseline,liang2018odin,lee2018mahalanobis,hendrycks2019corruptions,liu2020energy}. Adversarial-example research further provides evidence that high-confidence model behavior can be brittle under small or structured perturbations \citep{szegedy2014intriguing,goodfellow2015explaining}. Runtime perception-monitoring surveys connect these issues to safety-critical perception stacks; action authorization still calls for a decision layer that can block, modify, or escalate a proposed action \citep{schotschneider2025runtime}.

In Physical AI, this distinction is crucial. A model may be confident that an action follows from its learned representation, while the representation itself is wrong. A robot policy may confidently choose a trajectory based on a false free-space map. A drone may confidently continue navigation after subtle sensor drift. A VLA model may confidently interpret an instruction but ignore a workspace-specific safety rule. In each case, the model's confidence is a property of its internal computation, not a guarantee that the physical action is authorized \citep{guo2017calibration,ovadia2019uncertainty,matos2024sensorfailures,soh2026actionhallucination,ravichandran2025roboguard}.

This distinction matters for runtime authorization. Softmax or token confidence can indicate an internal model preference, but not physical safety or compliance with operational rules. Uncertainty estimates can indicate possible unreliability, but not the correct fallback action. OOD detectors can flag input mismatch with the training distribution, but not whether a specific proposed action is authorized. Task-success benchmarks can measure average performance on predefined scenarios, but not runtime safety under dynamic, evolving, or deployment-specific edge cases \citep{guo2017calibration,gal2016dropout,kendall2017uncertainties,lakshminarayanan2017ensembles,hendrycks2017baseline,liang2018odin,lee2018mahalanobis,hendrycks2019corruptions,ovadia2019uncertainty,liu2020energy,son2025subtlerisks,lu2025isbench}.

The guardrail problem is an action authorization problem. A runtime system is often expected to decide what to do next: authorize, modify, defer, request human input, degrade gracefully, switch to a fallback controller, block the action, or log the event for evaluation \citep{seto1998simplex,hobbs2023runtimeassurance,konighofer2022runtime,ravichandran2025roboguard,kim2026modular}.

\section{Runtime Authority as an Independent Layer}

The safety-control literature provides important tools for constraining actions. Safe reinforcement learning formalizes optimization under constraints \citep{garcia2015safesrl,achiam2017cpo,ray2019safetygym,zhao2023statewise,wachi2024constraint,gu2024safereview}. Control barrier functions, safety filters, reachability analysis, and predictive safety filters formalize the idea that actions should remain inside a safe set \citep{berkenkamp2017safe,fisac2019bridging,wabersich2021predictive,ames2019cbf,hsu2024safetyfilter,garg2024cbf}. Runtime assurance and shielding architectures similarly propose monitored autonomy, in which an advanced controller operates when acceptable and a trusted backup or enforcement mechanism intervenes when safety conditions are violated \citep{seto1998simplex,alshiekh2018shielding,hobbs2023runtimeassurance,konighofer2022runtime}.

These approaches are essential, but they do not by themselves cover the emerging Physical AI problem. Many assume access to system dynamics, low-level control variables, verified safe sets, or a limited autonomy stack \citep{ames2019cbf,wabersich2021predictive,hobbs2023runtimeassurance,hsu2024safetyfilter,garg2024cbf}. By contrast, deployed Physical AI systems may incorporate black-box world models, heterogeneous hardware, learned planners, vendor-specific controllers, cloud-edge pipelines, and site-specific constraints. Runtime guardrails are therefore analyzed here as multi-dimensional interfaces across semantic, state, physical, spatial, operational, and audit evidence.

The distinction is architectural rather than merely terminological. Semantic guardrails evaluate language, intent, content policy, and misuse \citep{kang2025polyguard,liu2026agentdog}. Safety filters and control barrier functions constrain low-level actions with respect to a modeled safe set \citep{ames2019cbf,wabersich2021predictive,hsu2024safetyfilter,garg2024cbf}. Runtime assurance monitors an operational controller and may switch to a trusted fallback \citep{seto1998simplex,hobbs2023runtimeassurance,konighofer2022runtime}. \textit{Physical action authorization} is the interface between these ideas: it asks whether a proposed action from a learned, possibly black-box model should be executed in the current state, under the current constraints, with a decision that can be reconstructed later \citep{ravichandran2025roboguard,kim2026modular,schotschneider2025runtime}. This framing makes the guardrail layer broader than content moderation, less controller-specific than a single safety filter, and more directly tied to deployment evidence than offline evaluation alone.

\subsection{Why Existing Guardrails Do Not Transfer Directly}

Existing guardrail approaches do not transfer directly to Physical AI for four reasons. First, semantic validity is not physical validity: an instruction can be benign, coherent, and policy-compliant while still producing an infeasible or unsafe trajectory \citep{soh2026actionhallucination,ravichandran2025roboguard,kim2026modular}. Second, controller-level safety filters often assume known dynamics, explicit state variables, and well-defined safe sets, whereas black-box Physical AI systems may output plans, waypoints, code-like actions, latent actions, or natural-language-conditioned policies \citep{ames2019cbf,hsu2024safetyfilter,kim2024openvla,black2024pi0,pertsch2025fast}. Third, runtime assurance assumes that the monitor can evaluate the active controller against trusted safety conditions; in Physical AI, the monitor also has to assess whether the state estimate itself is reliable enough to support the proposed action \citep{hobbs2023runtimeassurance,konighofer2022runtime,matos2024sensorfailures,schotschneider2025runtime}. Fourth, operational deployments introduce fleet-specific rules, restricted spaces, payload constraints, human-workflow constraints, and audit requirements that are not captured by generic model refusal policies \citep{ntsb2019tempe,cadmv2023cruise,nhtsa2023tesla}.

\begin{figure*}[!tbp]
\centering
\resizebox{0.80\textwidth}{!}{%
\begin{tikzpicture}[
  node distance=0.16cm,
  layer/.style={draw=StateLine,fill=white,rounded corners=2mm,minimum width=9.25cm,minimum height=0.74cm,align=center,font=\sffamily\small},
  terminal/.style={draw=StateNavy,fill=StateNavy!4,rounded corners=2mm,minimum width=4.2cm,minimum height=0.86cm,align=center,font=\sffamily\small},
  auth/.style={draw=StateCyan,fill=StateNavy,text=white,rounded corners=2mm,minimum width=9.25cm,minimum height=0.84cm,align=center,font=\sffamily\bfseries\small},
  arr/.style={-{Latex[length=2.1mm]},draw=StateSlate,line width=0.65pt}
]
  \node[terminal] (model) {Black-box Physical AI model};
  \node[layer,below=0.30cm of model] (proposal) {Proposed action + state evidence};
  \node[layer,below=of proposal] (sem) {Semantic guardrails: intent, prompt injection, harmful requests, human intent};
  \node[layer,below=of sem] (state) {State-validity guardrails: sensor integrity, perception anomaly, state-estimate consistency};
  \node[layer,below=of state] (phys) {Physical feasibility guardrails: kinematics, dynamics, collisions, velocity, payload, workspace envelope};
  \node[layer,below=of phys] (spatial) {Spatial and operational guardrails: geofencing, restricted zones, fleet rules, operational policy};
  \node[auth,below=0.18cm of spatial] (runtime) {Runtime authority: authorize \quad modify \quad block \quad fallback \quad escalate \quad log};
  \node[layer,below=0.24cm of runtime] (validated) {Validated action + audit trace};
  \node[terminal,below=0.30cm of validated] (hardware) {Hardware and controllers};

  \draw[arr] (model) -- (proposal);
  \draw[arr] (proposal) -- (sem);
  \draw[arr] (spatial) -- (runtime);
  \draw[arr] (runtime) -- (validated);
  \draw[arr] (validated) -- (hardware);
\end{tikzpicture}
}
\caption{Layered runtime authority. Semantic, state, physical, and operational checks are composed before the validated action and audit trace reach hardware.}
\label{fig:guardrail-stack}
\end{figure*}
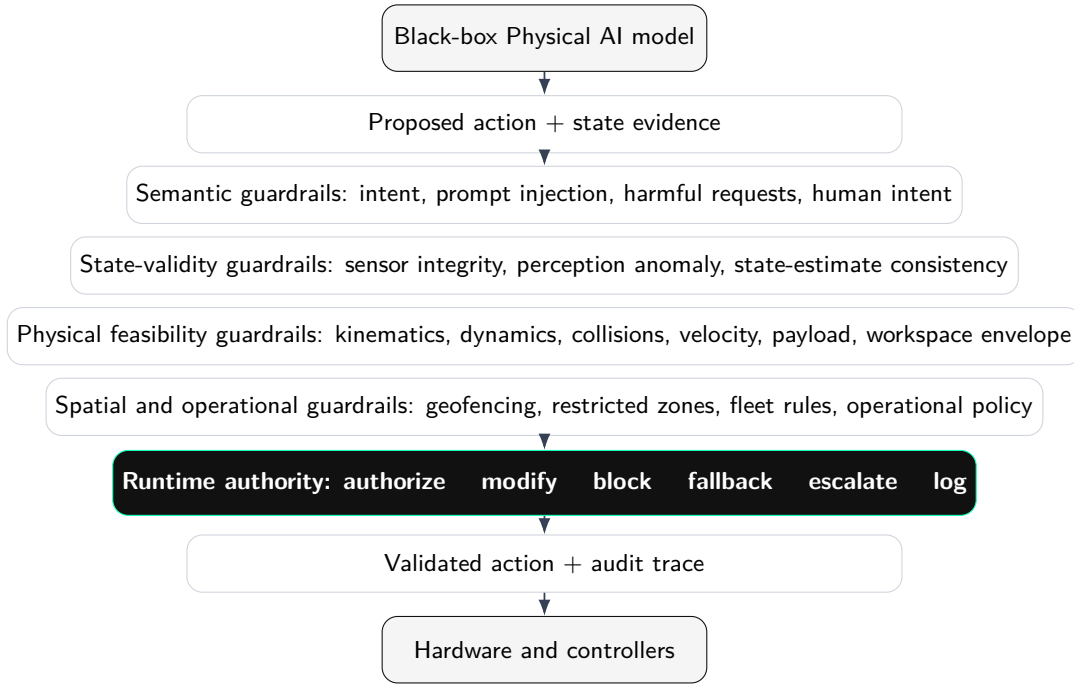

The implication is not that existing methods are insufficient individually; it is that their assumptions require explicit composition. Physical AI guardrails can be studied as interfaces that connect semantic policy, state validity, physical feasibility, spatial constraints, fallback behavior, and audit evidence around the same proposed action \citep{seto1998simplex,hobbs2023runtimeassurance,ravichandran2025roboguard,kim2026modular}. Figure~\ref{fig:guardrail-stack} illustrates this layered interpretation.

The architectural point is separation. A runtime guardrail ideally should not depend entirely on the same black-box model whose behavior it is evaluating \citep{seto1998simplex,hobbs2023runtimeassurance,konighofer2022runtime,ravichandran2025roboguard,kim2026modular}. This separation enables independent checks, deterministic rules, reproducible logs, and cross-model comparison. It also supports evaluation workflows in which the foundation model, hardware platform, task policy, or deployment environment can change while the action-authorization interface remains consistent.

Runtime guardrails may be hybrid. They may combine learned anomaly detection, formal constraints, geometric reasoning, model predictive checks, semantic policy engines, temporal logic, and classical control monitors \citep{hendrycks2017baseline,liu2020energy,wabersich2021predictive,ames2019cbf,hsu2024safetyfilter,schotschneider2025runtime,kang2025polyguard}. A key design principle is that the final authority over physical action should include inspectable constraints that are external to the generative model.

Recent guardrail work can be read as a sequence of increasingly deployment-oriented assumptions. RoboGuard argues that LLM-enabled robots require contextual safety rules and conflict resolution beyond generic content filtering, and reports large reductions in unsafe plan execution under its experimental setup \citep{ravichandran2025roboguard}. Modular guardrail work extends this point by arguing for multiple guardrail modules across action safety, decision safety, and human-centered safety in foundation-model-enabled robots \citep{kim2026modular}. These works are especially relevant to the Physical AI problem because they treat the model output as an object of mediation before robot execution.

Action hallucination and multimodal hallucination benchmarks sharpen the failure mechanism. Action hallucination work argues that generative VLA policies can emit physically invalid actions because the learned action distribution and feasible robot behavior are not necessarily aligned \citep{soh2026actionhallucination}. VideoHallu provides evidence that multimodal models can fail to identify physical, logical, or common-sense violations in synthetic video understanding tasks \citep{li2025videohallu}. Together, these works support the claim that plausibility, confidence, and perceptual fluency are not sufficient evidence for action authorization.

Poly-Guard and AgentDoG contribute a different but complementary direction. Poly-Guard benchmarks guardrail behavior across eight safety-critical domains and reports that optimized adversarial attacks remain effective against advanced guardrail models \citep{kang2025polyguard}. AgentDoG moves from binary safe/unsafe labels toward trajectory-level diagnosis of agentic risks and root causes \citep{liu2026agentdog}. These works make guardrails more contextual, policy-grounded, and auditable. Their limitation for Physical AI is not weakness; it is scope. Physical execution introduces additional evidence types: state validity, physical feasibility, spatial constraints, hardware limits, fallback behavior, and action-level audit records.

\section{Evaluation Under Dynamic Edge Cases}

Evaluation is central to Physical AI because safety risk accumulates through interaction among model uncertainty, environment change, human behavior, and hardware constraints. General benchmarks are useful, but static benchmarks can miss failure modes that only emerge through closed-loop behavior. Embodied safety benchmarks therefore increasingly evaluate agents across multi-step scenarios, hazardous instructions, adversarial prompts, object interactions, context-dependent constraints, physical-consistency errors, and trajectory-level risk diagnosis \citep{li2025videohallu,son2025subtlerisks,lu2025isbench,liu2026agentdog}.

For runtime guardrails, evaluation should measure more than task success. It should ask:

\begin{itemize}
  \item Did the system detect corrupted or insufficient state before action?
  \item Did it block physically invalid actions even when the model was confident?
  \item Did it enforce spatial, kinematic, and operational constraints consistently?
  \item Did it degrade gracefully under uncertainty?
  \item Did it produce auditable traces that explain why an action was authorized or rejected?
  \item Did the same guardrail policy generalize across models, hardware, and environments?
\end{itemize}

Rather than proposing a new benchmark or final protocol, this section identifies requirements that a comparative evaluation would need to satisfy. Existing benchmarks and simulator platforms illuminate important parts of the problem, but no single evaluation setup reviewed here covers the full runtime authorization pathway for Physical AI guardrails \citep{son2025subtlerisks,lu2025isbench,liu2026agentdog,nvidia2026isaacsim,makoviychuk2021isaacgym}. These requirements follow from the gaps between embodied safety benchmarks, simulator platforms, runtime assurance, uncertainty estimation, and guardrail datasets \citep{hobbs2023runtimeassurance,ovadia2019uncertainty,schotschneider2025runtime,kang2025polyguard}. Table~\ref{tab:evaluation-requirements} summarizes these requirements and the measurement problems that remain open.

\begin{table*}[!tbp]
\centering
\caption{Evaluation requirements derived from the reviewed benchmark, simulator, and assurance literatures.}
\label{tab:evaluation-requirements}
\footnotesize
\begin{tabularx}{\textwidth}{p{0.21\textwidth}p{0.34\textwidth}Y}
\toprule
\textbf{Requirement} & \textbf{Literature or formal anchor} & \textbf{Open measurement problem} \\
\midrule
State-reliability stress & Uncertainty, OOD, calibration, perception monitoring \citep{ovadia2019uncertainty,liu2020energy,matos2024sensorfailures,schotschneider2025runtime}. & When is state evidence sufficient for this action? \\
Invalid-action intervention & Action hallucination, robot jailbreaks, embodied guardrails \citep{soh2026actionhallucination,robey2024robopair,ravichandran2025roboguard}. & Necessary block or over-conservative interruption? \\
Operational constraints & Autonomy incidents and site constraints \citep{ntsb2019tempe,cadmv2023cruise,nhtsa2023tesla}. & How to compare heterogeneous deployment rules? \\
Pre-commit timing & Runtime assurance and physical commitment \citep{seto1998simplex,hobbs2023runtimeassurance}. & What counts as ``before execution'' across platforms? \\
Fallback correctness & Shielding and runtime enforcement \citep{alshiekh2018shielding,konighofer2022runtime,hobbs2023runtimeassurance}. & How to score stop, modify, and escalation? \\
Cross-model / platform & Cross-embodiment policies, simulators, robot datasets \citep{openx2023,octo2024,kim2024openvla,black2024pi0}. & How to compare without shared internals? \\
Audit completeness & Incident analysis and runtime records \citep{ntsb2019tempe,cadmv2023cruise,gm2024cruise}. & What is the minimal reconstructable record? \\
\bottomrule
\end{tabularx}
\end{table*}

These requirements can be instantiated by illustrative failure classes rather than treated as a fixed benchmark suite. Examples include an occluded obstacle in a mobile-robot aisle, a stale perception frame, a hallucinated manipulation affordance, a restricted-zone violation, a payload or velocity-limit violation, an adversarial instruction that passes semantic checks, a late intervention, and an unsafe fallback \citep{matos2024sensorfailures,soh2026actionhallucination,robey2024robopair,ravichandran2025roboguard,son2025subtlerisks,lu2025isbench}. The point is not that this list is exhaustive; it shows the range of evidence types that runtime guardrail evaluation should cover.

The requirements can be measured around the authorization record \(\xi_t\) defined in Equation~\eqref{eq:authorization-record}. A comparative study built around this record can compare models and guardrail layers without requiring that every system share the same internal architecture.

Let a comparative evaluation contain \(N\) authorization records \(\{\xi_i\}_{i=1}^{N}\). Let \(y_i=1\) denote that the proposed action is valid under the evaluation oracle and \(y_i=0\) denote that it is invalid. Let \(D_0=\{i:y_i=0\}\) be the invalid-action set, \(D_1=\{i:y_i=1\}\) the valid-action set, and \(\mathcal{I}=\{\mathrm{modify},\mathrm{block},\mathrm{escalate}\}\) the intervention decisions. Here \(|D|\) denotes set cardinality and \(\mathbb{I}[\cdot]\) denotes an indicator function. A primary safety metric is unsafe-action intervention rate:
\begin{equation}
\eqfit{\mathrm{UAIR}=\frac{1}{|D_0|}\sum_{i\in D_0}\mathbb{I}[\rho_i\in\mathcal{I}]}
\label{eq:unsafe-action-intervention-rate}
\end{equation}
The corresponding operational-cost metric is false block rate, \(\mathrm{FBR}=|D_1|^{-1}\sum_{i\in D_1}\mathbb{I}[\rho_i\in\mathcal{I}]\), which captures usability cost rather than direct safety prevention. For physical systems, timing is itself a safety variable. If \(t_i^{\mathrm{prop}}\) is the proposal time, \(t_i^{\rho}\) is the guardrail decision time, and \(t_i^{\mathrm{commit}}\) is the latest time before physical commitment, then pre-commit intervention rate is
\begin{equation}
\eqfit{\mathrm{PCIR}=\frac{1}{|D_0|}\sum_{i\in D_0}\mathbb{I}[\rho_i\in\mathcal{I}]\,\mathbb{I}[t_i^{\rho}<t_i^{\mathrm{commit}}]}
\label{eq:precommit-intervention-rate}
\end{equation}
Finally, guardrails should be evaluated not only by binary decisions but also by residual violation severity. If \(a_i^{\mathrm{out}}\) is the action after guardrail intervention, \(K\) is the number of evaluated constraints, and each normalized constraint is written as \(\tilde{c}_k(a_i^{\mathrm{out}},s_i)\leq 0\), a residual violation score is
\begin{equation}
\eqfit{\mathrm{RVS}=\frac{1}{NK}\sum_{i=1}^{N}\sum_{k=1}^{K}\max\{0,\tilde{c}_k(a_i^{\mathrm{out}},s_i)\}}
\label{eq:residual-violation-score}
\end{equation}
Equations~\eqref{eq:unsafe-action-intervention-rate}--\eqref{eq:residual-violation-score} are not meant to exhaust evaluation; they define a minimum quantitative core for comparing guardrails. Table~\ref{tab:evaluation-schema} groups the quantitative core with the additional qualitative checks needed for deployment review.

These metrics also have clear boundary cases. \(\mathrm{UAIR}=1\) means every invalid proposed action received an intervention, while \(\mathrm{UAIR}=0\) means none did. \(\mathrm{FBR}=0\) means valid actions were not unnecessarily interrupted; a high \(\mathrm{FBR}\) indicates an over-conservative guardrail. \(\mathrm{PCIR}=\mathrm{UAIR}\) when all interventions arrive before physical commitment, whereas \(\mathrm{PCIR}<\mathrm{UAIR}\) indicates that some interventions are too late to prevent execution. \(\mathrm{RVS}=0\) means the post-guardrail action satisfies all measured constraints; larger values indicate residual violation magnitude. If \(D_0\) or \(D_1\) is empty, the corresponding rate should be reported as not applicable rather than zero.

\begin{table*}[!tbp]
\centering
\caption{Compact metric families for Physical AI runtime guardrail evaluation.}
\label{tab:evaluation-schema}
\footnotesize
\begin{tabularx}{\textwidth}{p{0.24\textwidth}p{0.32\textwidth}Y}
\toprule
\textbf{Metric family} & \textbf{Operational definition} & \textbf{Why it matters} \\
\midrule
Intervention effectiveness & UAIR: invalid proposals interrupted. & Catches unsafe proposals. \\
Operational over-blocking & FBR: valid proposals interrupted. & Captures usability cost. \\
Pre-commit timing & PCIR and decision latency. & Detects late interventions. \\
State-validity sensitivity & Corrupt, stale, shifted, or inconsistent state evidence. & Separates action safety from state trust. \\
Constraint coverage & Semantic, state, physical, temporal, spatial, operational constraints. & Tests breadth beyond content filters. \\
Residual violation severity & RVS after intervention. & Measures remaining violation magnitude. \\
Fallback correctness & Block, modify, or escalation leads to safe behavior. & Avoids unsafe recovery. \\
Audit completeness & Evidence, constraints, versions, reason codes. & Supports review and reproducibility. \\
\bottomrule
\end{tabularx}
\end{table*}

Simulation environments are therefore important, but their role should be stated precisely. They can generate edge cases, synthetic sensor data, software- or hardware-in-the-loop tests, and repeatable task conditions \citep{nvidia2026isaacsim,makoviychuk2021isaacgym,dosovitskiy2017carla,shah2018airsim,savva2019habitat,xiang2020sapien}; they do not by themselves decide whether a black-box model's proposed action should be authorized at runtime. Table~\ref{tab:simulation-platforms} compares representative simulator families and the corresponding action-authorization gap. The right column is an interpretation of the authorization gap, not a claim made by the simulator papers themselves.

\begin{table*}[!tbp]
\centering
\caption{Representative simulation platforms and their relationship to runtime action authorization.}
\label{tab:simulation-platforms}
\footnotesize
\begin{tabularx}{\textwidth}{p{0.20\textwidth}p{0.37\textwidth}Y}
\toprule
\textbf{Simulator / platform family} & \textbf{Primary contribution for evaluation} & \textbf{Remaining gap for guardrails} \\
\midrule
NVIDIA Isaac Sim / Lab / Gym & Physically based simulation, synthetic data, GPU robot learning \citep{nvidia2026isaacsim,mittal2025isaaclab,makoviychuk2021isaacgym}. & Scales testing; does not authorize actions. \\
Newton physics engine & GPU physics for robot learning on Warp/OpenUSD \citep{nvidia2025newton}. & Improves physics substrate, not runtime policy. \\
MuJoCo / Gazebo & Dynamics, contact-rich control, multi-robot experimentation \citep{todorov2012mujoco,koenig2004gazebo}. & Controller tests are not action authorization. \\
CARLA / AirSim & Driving and aerial simulation with sensors and environments \citep{dosovitskiy2017carla,shah2018airsim}. & Scenario tests lack deployment audit authority. \\
Habitat / AI2-THOR & Embodied navigation and indoor interaction \citep{savva2019habitat,kolve2017ai2thor}. & Task completion is not physical validity. \\
SAPIEN / ManiSkill2 & Manipulation and articulated-object benchmarks \citep{xiang2020sapien,gu2023maniskill2}. & Needs feasibility, fallback, audit decisions. \\
\bottomrule
\end{tabularx}
\end{table*}

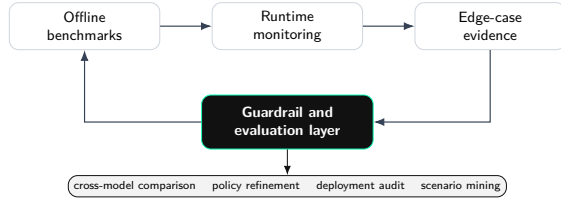
\begin{figure}[!tbp]
\centering
\resizebox{0.86\linewidth}{!}{%
\begin{tikzpicture}[
  box/.style={draw=StateLine,fill=white,rounded corners=2mm,minimum width=3.2cm,minimum height=1.0cm,align=center,font=\sffamily\small},
  gap/.style={draw=StateCyan,fill=StateNavy,text=white,rounded corners=2mm,minimum width=3.65cm,minimum height=1.15cm,align=center,font=\sffamily\bfseries\small},
  arr/.style={-{Latex[length=2.3mm]},draw=StateSlate,line width=0.7pt}
]
  \node[box] (offline) {Offline\\benchmarks};
  \node[box,right=1.1cm of offline] (runtime) {Runtime\\monitoring};
  \node[box,right=1.1cm of runtime] (logs) {Edge-case\\evidence};
  \node[gap,below=0.95cm of runtime] (platform) {Guardrail and\\evaluation layer};

  \draw[arr] (offline) -- (runtime);
  \draw[arr] (runtime) -- (logs);
  \draw[arr] (logs.south) |- (platform.east);
  \draw[arr] (platform.west) -| (offline.south);

  \node[draw=StateNavy,fill=StateNavy!5,rounded corners=2mm,below=0.55cm of platform,minimum width=8.0cm,align=center,font=\sffamily\scriptsize] (evidence)
  {cross-model comparison \quad policy refinement \quad deployment audit \quad scenario mining};
  \draw[-{Latex[length=2mm]},draw=StateNavy,line width=0.6pt] (platform) -- (evidence);
\end{tikzpicture}
}
\caption{Continuous evaluation loop. Offline benchmarks and runtime logs can feed the same guardrail evaluation process through authorization records.}
\label{fig:evaluation-loop}
\end{figure}

Together, world models and simulation environments support a continuous-evidence workflow: offline scenario generation, runtime monitoring, edge-case discovery, and guardrail refinement \citep{li2025worldmodels,hou2026worldrobot,nvidia2026physicalaidatafactory,schotschneider2025runtime}. Figure~\ref{fig:evaluation-loop} summarizes how benchmark evidence and deployment evidence can inform the same guardrail evaluation process.

Offline tests can characterize model behavior before deployment, while runtime logs can reveal new edge cases during operation \citep{son2025subtlerisks,lu2025isbench,liu2026agentdog,schotschneider2025runtime}. As Physical AI systems become more general and more frequently updated, evaluation is better treated as a continuing deployment concern rather than a one-time benchmark result.

\section{Synthesis: Action-Authorization Gap and Open Questions}

The literature points to an action-authorization gap across model-centric, control-centric, and safety-centric work. Model-centric work focuses on capabilities, datasets, architectures, and training \citep{openx2023,octo2024,kim2024openvla,black2024pi0,pertsch2025fast}. Robotics and control work focuses on hardware, dynamics, and task performance \citep{ames2019cbf,wabersich2021predictive,hsu2024safetyfilter,garg2024cbf}. Safety research contributes formal methods, uncertainty quantification, runtime monitoring, and evaluation protocols \citep{katz2017reluplex,gehr2018ai2,singh2019abstract,ovadia2019uncertainty,hobbs2023runtimeassurance,schotschneider2025runtime}. The open problem is how to connect these perspectives into a repeatable method for authorizing physical action under uncertainty.

\begin{statebox}{Current Literature Gap}
Existing surveys and technical works cover VLA models, world models, safety filters, runtime assurance, OOD detection, and embodied robot safety \citep{xu2024robotics,ma2024vlasurvey,sapkota2025vla,li2025worldmodels,hou2026worldrobot,hobbs2023runtimeassurance,gu2024safereview,schotschneider2025runtime,kim2026modular}. A remaining assurance category is model-independent runtime authorization for translating heterogeneous Physical AI outputs into validated, constrained, auditable physical actions. The gap is not a lack of safety techniques; it is the limited availability of a common authorization boundary that composes these techniques around proposed physical actions.
\end{statebox}

\subsection{Fragmented Safety Assumptions}

Classical control methods often assume well-specified dynamics and safe sets \citep{ames2019cbf,wabersich2021predictive,hsu2024safetyfilter,garg2024cbf}. Foundation-model works often evaluate success rates and generalization \citep{openx2023,octo2024,kim2024openvla,black2024pi0,pertsch2025fast}. LLM and multimodal safety research often focuses on harmful instructions, hallucination, jailbreaks, and policy compliance \citep{kang2025polyguard,li2025videohallu,liu2026agentdog,robey2024robopair}. Physical AI deployments bring these concerns together: semantic policy, state reliability, and deterministic physical constraints.

\subsection{Limited Model-Independent Runtime Authority}

Many safety interventions are embedded inside a specific model, robot, or task \citep{ravichandran2025roboguard,kim2026modular,liu2026agentdog}. A model-independent interface for proposed actions, state evidence, operational constraints, and authorization decisions would make it easier to compare systems without forcing every model or robot into the same internal representation. This supports research reproducibility, cross-model evaluation, system comparison, and heterogeneous deployments.

\subsection{Insufficient Treatment of Silent Failures}

Terms such as hallucination, distribution shift, OOD detection, perception error, and adversarial attack describe parts of the problem \citep{hendrycks2017baseline,ovadia2019uncertainty,liu2020energy,matos2024sensorfailures,soh2026actionhallucination,li2025videohallu,argota2026particles}. Silent failure provides a deployment-oriented framing: the system is still running, still confident, and still acting, but the action is grounded in an invalid state. This framing connects model-level and perception-level errors to physical consequences.

\subsection{Limited Auditability for Physical Action}

Physical AI systems are easier to evaluate when records explain why actions were allowed, modified, blocked, or escalated. Auditability supports scientific reproducibility, benchmarking, incident analysis, operational accountability, and regulatory readiness, especially when failures involve interacting perception, planning, control, and organizational factors \citep{ntsb2019tempe,cadmv2023cruise,nhtsa2023tesla,gm2024cruise,hobbs2023runtimeassurance}. Without structured traces, it becomes difficult to determine whether a failure came from perception, model reasoning, guardrail configuration, controller behavior, or the environment.

\subsection{Open Questions}

The surveyed literature motivates several open questions for runtime action authorization.

\begin{enumerate}
  \item \textbf{Action interface.} What is the right abstraction for a proposed physical action across drones, autonomous mobile robots, vehicles, manipulators, and humanoids?
  \item \textbf{State reliability.} How can runtime systems quantify whether the current world representation is reliable enough for a specific action?
  \item \textbf{Constraint composition.} How should semantic, spatial, kinematic, operational, and safety constraints be combined without creating brittle rule systems?
  \item \textbf{Guardrail evaluation.} What comparative evaluation methods can measure whether a guardrail layer reduces or detects silent failures beyond ordinary task completion?
  \item \textbf{Cross-platform learning.} How can deployment logs and edge cases improve guardrail policies while preserving model independence and auditability?
  \item \textbf{Runtime governance.} When should a system block, modify, defer, or escalate a proposed action, and how should that decision be explained?
\end{enumerate}

This synthesis complements advances in model training and control. Better world models may reduce some errors, yet open-world conditions make it difficult to treat any model as perfect \citep{li2025worldmodels,hou2026worldrobot,maes2026leworldmodel,chen2026abotphysworld}. Better controllers can reduce some violations, while upstream state validity and task intent call for separate evidence \citep{ames2019cbf,wabersich2021predictive,hsu2024safetyfilter,matos2024sensorfailures}. Runtime guardrails address the intermediate event: a black-box Physical AI system proposes an action, and the surrounding system evaluates whether that action is allowed.

\subsection{Falsifiable Implications}

The main inference is falsifiable. It would be weakened if future Physical AI systems routinely exposed certified action representations, reliable state-validity evidence, formally composable constraints, and auditable authorization decisions as part of the model or controller interface. It would also be weakened if empirical benchmarks showed that model-internal confidence, low-level safety filters, or semantic guardrails alone consistently reduce or prevent silent physical-action failures across heterogeneous robots, vehicles, drones, environments, and operational policies. Conversely, repeated findings that plausible model outputs benefit from external state, feasibility, spatial, temporal, or audit checks would strengthen the case for an independent runtime authorization layer.

\section{Assurance Implications and Minimal Event Schema}

A technical implication is that guardrail evaluation should measure intervention quality, not only task success \citep{son2025subtlerisks,lu2025isbench,liu2026agentdog}. Researchers can treat the authorization event as a concrete unit of analysis: whether unsafe, invalid, or poorly grounded actions are detected, modified, blocked, or escalated before execution. System architects can separate this interface from model training, low-level control, and hardware-specific safety mechanisms \citep{seto1998simplex,hobbs2023runtimeassurance,kim2026modular}. Deployment reviews can ask whether the system produces independent evidence about why actions were allowed or rejected under uncertainty \citep{ntsb2019tempe,cadmv2023cruise,gm2024cruise}.

The contribution is a structured map of the gap between model capability and physical-action assurance: the failure mode, the relevant literatures, the minimum formal interface, evaluation requirements, representative metric families, and the evidence needed to compare guardrail layers across platforms. It is not a new standard or a complete guardrail design.

\subsection{Minimal Authorization Event Schema}
\label{sec:minimal-authorization-schema}

A minimal unit for comparing runtime guardrails is the authorization event. For each proposed physical action, a structured record can be defined independently of the internal representation of the policy, world model, simulator, or controller. Such a record does not prescribe a software API or implementation standard; it defines the information needed to evaluate whether an action was authorized, modified, blocked, escalated, or routed to fallback.

\begin{table*}[!tbp]
\centering
\caption{Minimal authorization event schema for comparing runtime guardrails.}
\label{tab:minimal-authorization-schema}
\footnotesize
\begin{tabularx}{\textwidth}{p{0.22\textwidth}p{0.36\textwidth}Y}
\toprule
\textbf{Field} & \textbf{Meaning} & \textbf{Why it matters} \\
\midrule
Observation context & Sensor inputs, history, timestamp, platform context. & What evidence was available. \\
Proposed action & Command, waypoint, plan, tool action, or trajectory fragment. & Separates proposal from execution. \\
State estimate & Objects, agents, maps, robot state, environment state. & States what authorization depends on. \\
State-validity evidence & Sensor health, latency, uncertainty, OOD, consistency flags. & Separates unsafe action from unreliable state. \\
Active constraints & Physical, semantic, spatial, temporal, operational, policy constraints. & Defines the evaluation boundary. \\
Authorization decision & Authorize, modify, block, escalate, or fallback. & Records runtime judgment. \\
Fallback or modification & Replacement action, safe stop, backup controller, human escalation. & Captures recovery semantics. \\
Timing evidence & Proposal, decision, commitment times, latency. & Checks pre-commit intervention. \\
Audit trace & Versions, constraint IDs, reason codes, evidence pointers. & Supports incident analysis and comparison. \\
\bottomrule
\end{tabularx}
\end{table*}

The schema is deliberately limited to fields that recur across platforms. Different platforms may instantiate the fields differently: an aerial system may emphasize geofences, energy margins, and GNSS integrity; a mobile robot may emphasize clearance, occupancy maps, and human proximity; a manipulator may emphasize contact constraints, payload, workspace limits, and grasp feasibility. The common point is that the unit of comparison is not the model architecture or the controller alone, but the authorization event that links a proposed action to state evidence, active constraints, runtime decision, fallback, and auditability.

\section{Limitations and Threats to Validity}

The analysis is limited to runtime guardrails and action authorization for black-box Physical AI systems. It does not attempt to survey the entire Physical AI literature, nor does it provide a complete treatment of robotics standards, simulation infrastructure, certification processes, or general AI ethics. The source selection is centered on the action-authorization pathway; therefore, some adjacent literatures are represented only when they directly clarify the connection between model output, state evidence, physical constraints, and execution.

Several limitations follow from this scope. First, the formalization is intentionally minimal and abstracts away many hardware-specific dynamics, perception-stack details, and organizational deployment processes. Second, the autonomous-vehicle incidents discussed above are used as operational analogues for assurance failures in physical autonomy; they should not be read as direct examples of VLA or world-model foundation systems. Third, the review derives evaluation requirements and metric families but does not introduce a new benchmark or empirical evaluation of a guardrail implementation. Fourth, runtime guardrails are not presented as a complete safety solution. They are one assurance layer among model design, perception robustness, classical control, hardware redundancy, human oversight, and organizational safety processes.

There are also technical boundary conditions. Runtime authorization cannot guarantee safety when observability is poor, constraints are incomplete, the fallback behavior is underspecified, or the physical environment is adversarial. Formal guarantees from safety filters, CBFs, or runtime assurance can weaken when state estimates are stale, when the dynamics model is wrong, or when the action representation exposed by the learned policy is not the representation assumed by the safety proof \citep{ames2019cbf,wabersich2021predictive,hobbs2023runtimeassurance,ovadia2019uncertainty,matos2024sensorfailures}. Authorization latency can itself become hazardous: a correct block that arrives after physical commitment is operationally insufficient. Finally, fallback is not automatically safe; safe stop, human escalation, and backup control each require their own validity conditions.

These limitations are also useful boundary conditions. A runtime guardrail layer should not replace certified control, verified hardware limits, operator training, incident response, or regulatory compliance. Its role is narrower and more specific: to provide a model-independent authorization boundary where proposed physical actions can be checked, modified, blocked, escalated, and logged before they become hardware commitments.

\textbf{Selection validity.} The source selection is curated around runtime action authorization and therefore may underrepresent adjacent work whose relevance is indirect, such as broader robotics standards, human factors, or simulation infrastructure.

\textbf{Construct validity.} The term ``silent failure'' groups several mechanisms: hallucination, OOD behavior, sensor drift, state-estimation error, and invalid affordance inference. This grouping is useful for deployment analysis, but each mechanism may call for different instrumentation and mitigation.

\textbf{External validity.} Physical AI deployments vary across robots, vehicles, drones, factories, warehouses, hospitals, roads, and homes. A guardrail taxonomy that is useful across domains still needs domain-specific constraints, fallback policies, and evidence thresholds.

\textbf{Temporal validity.} The literature is moving quickly. New VLA models, world models, robot benchmarks, and guardrail datasets may narrow parts of the gap identified here. The claim should therefore be read as an argument about an assurance category, not a fixed statement about any single model generation.

\textbf{Evidence validity.} The autonomous-vehicle incidents discussed above provide operational analogues for runtime monitoring and boundary failures. They are not direct evidence about current VLA or world-model-based robotic systems, and they are used only as examples of how physical autonomy failures can create safety and regulatory consequences.

\section{Conclusion}

The main conclusion is that Physical AI safety is not only a model-training problem, a controller problem, or a simulation problem. It is also an authorization problem. Before a learned system's proposed action becomes a physical commitment, the surrounding autonomy stack should provide independent evidence that the state is valid, the action is feasible, operational constraints are satisfied, fallback is available, and the decision is auditable.

The critical failure mode is often silent: a system acts confidently on a corrupted or incomplete world state. The literature provides many ingredients for addressing this problem, including safe control, runtime assurance, shielding, uncertainty estimation, OOD detection, embodied safety benchmarks, neural-network verification, simulation infrastructure, and robot-specific guardrails. One unresolved task is their integration into a runtime authority that evaluates and records physical-action decisions before execution.

Three implications follow. Researchers should evaluate authorization events, not only task success. System builders should separate model proposal from runtime authority, especially when models, hardware, policies, and environments change independently. Safety teams and regulators can use reconstructable action records to examine why actions were authorized, modified, blocked, or escalated under uncertainty.

Runtime guardrails can therefore be studied as assurance mechanisms for the transition from AI that predicts the world to AI that acts in it.

\bibliographystyle{plainnat}
\bibliography{references}

\end{document}